\documentclass{article}

\usepackage{acro}
\usepackage{spconf}
\usepackage{subfig}
\usepackage{amsmath}
\usepackage{caption} 
\usepackage{amsfonts}
\usepackage{cleveref}
\usepackage{multirow}
\usepackage{graphicx}
\usepackage[table]{xcolor}
\usepackage[normalem]{ulem}
\usepackage[export]{adjustbox}
\useunder{\uline}{\ul}{}

\makeatletter
\newcommand{\ie}{\emph{i.e.}\@ifnextchar.{\!\@gobble}{}}
\newcommand{\eg}{\emph{e.g.}\@ifnextchar.{\!\@gobble}{}}
\newcommand{\etc}{etc\@ifnextchar.{}{.\@}}
\makeatother

\useunder{\uline}{\ul}{}

\def\Pcollision{\ensuremath{P_\text{collision}}}
\def\LossNHD{\ensuremath{L_\text{NHD}}}
\def\LossPseudo{\ensuremath{L_\text{pseudo}}}
\def\LossAttention{\ensuremath{L_\text{att}}}
\def\LambdaPseudo{\ensuremath{\lambda_\text{pseudo}}}
\def\LambdaAttention{\ensuremath{\lambda_\text{att}}}
\def\LossCode{\ensuremath{L_\text{code}}}
\def\LossDEC{\ensuremath{L_\text{DEC}}}
\def\Wpseudo{\ensuremath{W^{\text{(pseudo)}}}}

\DeclareAcronym{mAP}{short = mAP, long = mean Average Precision}
\DeclareAcronym{name}{short = CS3H, long = Collision-Resistant Single-Pass Self-Supervised Semantic Hashing}

\title{Collision-Resistant Single-Pass Method for Unsupervised Fine-Grained Image Hashing}
\name{Anh-Kiet Duong$^{1,2}$\thanks{This work was supported by the ANR project ExcelLR (ANR-21-EXES-0010) and the L3i Laboratory computing resources.}, Petra Gomez-Krämer$^{1}$, Jean-Michel Carozza$^{2}$}
\address{$^{1}$ L3i Laboratory, La Rochelle University, 17042 La Rochelle Cedex 1, France\\
$^{2}$ LIENSs Laboratory, La Rochelle University, 17042 La Rochelle Cedex 1, France \\
\tt\small \{anh.duong,petra.gomez,jean-michel.carozza\}@univ-lr.fr
}

\begin{document}
\maketitle

\begin{abstract}
Unsupervised fine-grained image hashing aims to learn compact binary codes 
that preserve subtle visual differences among highly similar instances without 
manual annotations. However, most existing methods neglect collision resistance, 
leading to identical hash codes for slightly semantically different samples. 
In this paper, we propose \ac{name}, a collision-resistant framework that 
directly optimizes Hamming-space similarity via a single-pass normalized 
Hamming distance loss to produce well-separated binary representations. 
We further introduce a collision-sensitive attention module to emphasize rare 
and discriminative local patterns, reducing hash collisions and improving 
fine-grained discrimination. Experiments on multiple benchmarks show 
that \ac{name} consistently outperforms state-of-the-art methods in retrieval 
accuracy while achieving superior collision resistance with minimal 
computational overhead.
\end{abstract}

\section{Introduction}
\label{sec:intro}

Fine-grained image retrieval is challenging due to subtle visual differences among highly similar instances. Many existing methods rely on high-dimensional features, resulting in high computational and memory costs at scale~\cite{luo2023survey}. Deep hashing mitigates this by encoding images into compact binary codes for efficient Hamming-space search. However, recent fine-grained hashing methods often depend on labeled data or multi-view training and largely ignore collision resistance, leading to degraded retrieval performance. Moreover, the only unsupervised fine-grained hashing approach~\cite{hu2024asymmetric} does not release code, leaving no reproducible baseline.

In this work, we propose \ac{name}, a collision-resistant single-pass framework for unsupervised fine-grained image hashing. By directly optimizing in the Hamming space, our method avoids multi-view contrastive learning and generates distinctive binary codes. The framework integrates a normalized Hamming distance loss and a collision-sensitive attention module focusing on rare discriminative regions, enabling efficient end-to-end training and establishing the first reproducible baseline for this task.
\begin{itemize}
\setlength{\itemsep}{0ex}
\setlength{\parskip}{0ex}
\setlength{\parsep}{0ex}
    \item We revisit collision resistance from cryptography and adapt it for deep image hashing.
    \item We propose a single-pass framework that directly optimizes similarity in the Hamming space, improving retrieval performance while maintaining low complexity.
    \item We introduce a collision-sensitive attention mechanism to emphasize rare and discriminative patterns.
\end{itemize}

The rest of this paper is organized as follows. \Cref{sec:related} reviews related work. \Cref{sec:method} presents the proposed method. \Cref{sec:experiments} reports experimental results, and \cref{sec:conclusion} concludes the paper.

\section{Related work}
\label{sec:related}

Unsupervised fine-grained image hashing is a relatively new topic first introduced by~\cite{hu2024asymmetric}. Existing unsupervised hashing methods such as~\cite{qiu2021unsupervised, wang2022contrastive, ng2023unsupervised} have demonstrated competitive performance on generic image datasets like NUS-WIDE. However, these methods show a noticeable performance drop when applied to fine-grained datasets. This performance gap highlights the need for new approaches specifically designed to handle the challenges of fine-grained image retrieval. Quantization-based approaches~\cite{jang2021self, wang2022contrastive} also increase complexity due to exponential clustering. Moreover, most related methods suffer from high computational complexity due to multi-pass training and remain vulnerable to hash collisions, which are particularly problematic for hashing.

\section{Methodology}
\label{sec:method}
\begin{figure*}[ht]
    \centering
    \includegraphics[width=0.9\textwidth]{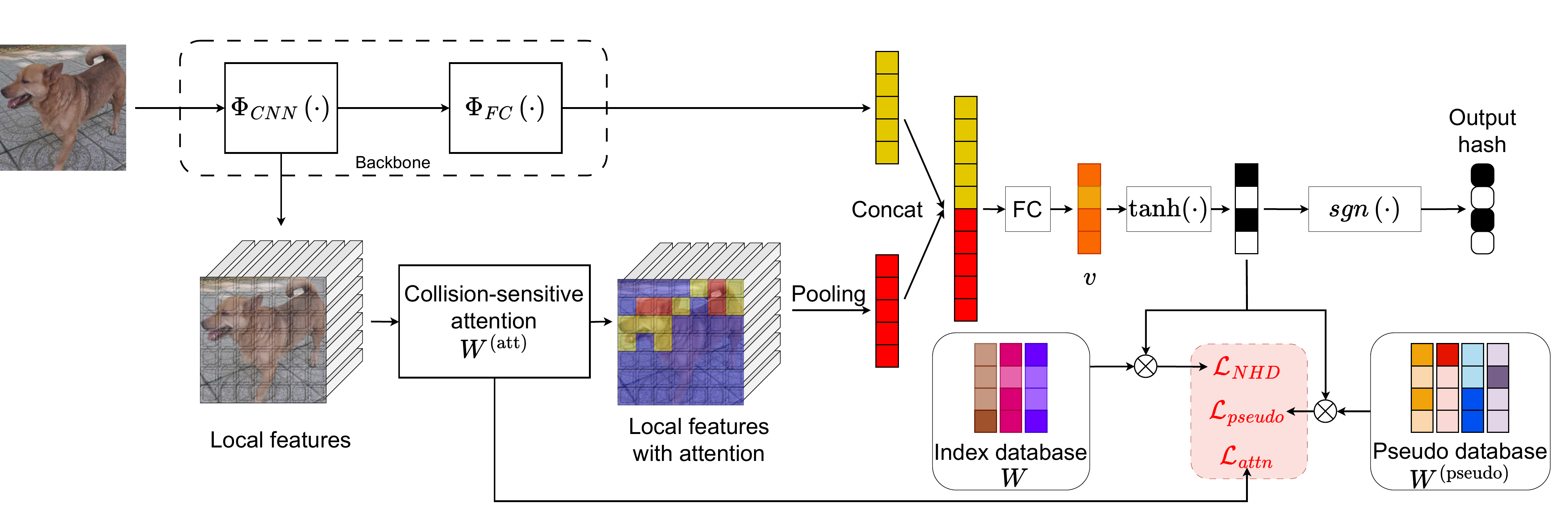}
    \caption{Overview of the proposed framework. The network extracts global and local features to produce hash codes optimized under three objectives: a normalized Hamming distance loss for direct optimization in the Hamming space, a pseudo labeling module for semantic consistency, and a collision-sensitive attention module that emphasizes distinctive, collision-resistant cues.}
    \label{fig:framework}
\end{figure*}

Given a dataset $\mathcal{D}=\left\{ x_i \right\}_{i=1}^{N}$ of size $N$, a hash function is defined as $\mathcal{H}:x \mapsto \left\{0, 1\right\}^{l}$, where $l$ denotes the hash code length. In deep image hashing, the learned codes are required to be query-friendly, meaning that the Hamming distance between $\mathcal{H}\left(x_i\right)$ and $\mathcal{H}\left(x_j\right)$ is small for visually similar samples. Following common practice \cite{hu2024asymmetric}, we use $\tanh(\cdot)$ as a soft quantization and equivalently define $\mathcal{H}:x \mapsto \left\{-1, 1\right\}^{l}$, where $-1$ is stored as bit $0$ in practice. Based on this formulation, our framework learns compact and collision-resistant hash codes in a single forward pass by jointly optimizing a normalized Hamming distance loss, pseudo-label consistency, and a collision-sensitive attention mechanism, as illustrated in \cref{fig:framework}.

\subsection{Hash collision}
\label{sec:collision}

In cryptography~\cite{menezes2018handbook}, hash functions are expected to satisfy collision resistance, pre-image resistance, and second pre-image resistance. In deep image hashing, pre-image related properties depend heavily on the backbone and threat model, while collision resistance can be explicitly quantified. We define a collision indicator:
\begin{equation}\label{eq:collision}
\resizebox{0.44\textwidth}{!}{$
\Pcollision
= \frac{
    \sum_{i=1}^{N} \sum_{j=i+1}^{N}
    \mathrm{col}\!\left(
        \mathcal{H}(x_i), \mathcal{H}(x_j)
    \right)
}{\binom{N}{2}},
\quad
\mathrm{col}(a,b)=
\begin{cases}
1, & \text{if } a=b,\\
0, & \text{if } a\ne b.
\end{cases}
$ .}
\end{equation}
A smaller $\Pcollision$ indicates stronger collision resistance, with the ideal case $\Pcollision=2^{-l}$ corresponding to a random $l$-bit hash function. Despite its importance, existing unsupervised hashing methods mainly optimize retrieval accuracy and largely neglect collision resistance~\cite{wang2022contrastive, ng2023unsupervised, hu2024asymmetric}. In fine-grained retrieval, this property is crucial for distinguishing individual instances within the same category, motivating us to explicitly design a collision-resistant hashing framework.

\subsection{Normalized Hamming distance loss}
\label{sec:loss}

Recent works often adopt clustering-based hashing~\cite{jang2021self, wang2022contrastive}, where the $l$-bit hash is divided into $M$ codebooks with $2^{l/M}\!\cdot\!M$ centroids in total. While this design enhances local diversity, it substantially increases memory and computational complexity. To keep the framework lightweight, we instead adopt a simple sign quantization $\operatorname{sgn}(\cdot)$~\cite{cao2017hashnet}. A fully connected layer at the end of the model outputs a continuous vector $v \in \mathbb{R}^l$, and the final hash code is obtained as:
\begin{equation}
    u = \operatorname{sgn} \left(  v \right),
    \operatorname{sgn}(x) = \begin{cases} 
    1, & \text{if } x \geq 0, \\
    -1, & \text{if } x < 0.
    \end{cases}
\end{equation}

Since $\operatorname{sgn}(\cdot)$ is non-differentiable, it is only applied during inference. During training, we replace it with $\tanh(\cdot)$, which smoothly saturates to $[-1, 1]$ and allows gradient backpropagation. The contrastive objective from~\cite{hu2024asymmetric} is defined as:
\begin{equation}
\resizebox{0.44\textwidth}{!}{ 
${{L}_{1}}=-\log \frac{{{e}^{s \left( \tanh \left( v_{i}^{\left( a \right)} \right)\cdot \tanh \left( v_{i}^{\left( p \right) } \right) \right)}}}{{{e}^{s \left( \tanh \left( v_{i}^{\left( a \right)} \right)\cdot \tanh \left( v_{i}^{\left( p \right)} \right) \right)}} + \sum\limits_{j=1}^{2{{N}_{B}}-1}{{{e}^{s \left( \tanh \left( v_{i}^{\left( a \right)} \right)\cdot \tanh \left( v_{i}^{\left( n \right)} \right) \right)}}}}$
}.
\end{equation}
Here, $N_B$ denotes the batch size and $s$ the scale parameter. Each batch contains $3N_B$ images, including anchor $\left(a\right)$, positive $\left(p\right)$, and negative $\left(n\right)$ views. In practice, large batch sizes are required to provide sufficient negatives, leading to high memory consumption due to multi-view processing. To address this limitation, we propose to reformulate $L_1$ into a single-view classification-based objective as:
\begin{equation}
\label{eq:loss_single}
    L_{2} = -\log 
    \frac{
        e^{s \left( \tanh(W_{i}) \cdot \tanh(v_{i}) \right)}
    }{
        \sum_{j=1}^{N} e^{s \left( \tanh(W_{j}) \cdot \tanh(v_{i}) \right)}
    }.
\end{equation}
Here, $W \in \mathbb{R}^{N\times l}$ is a learnable memory bank storing latent features for all $N$ samples, where $W_i$ corresponds to the $i$-th image. Unlike supervised hashing~\cite{jiang2018asymmetric}, which relies on label-based classification with fixed binary codes and is not applicable in the unsupervised setting, we perform instance-level classification using a continuous memory bank jointly optimized with the network. This design removes multi-view overhead, provides more negatives, improves convergence and collision resistance.

However, dot-product objectives in $L_1$ and $L_2$ do not directly optimize similarity in the Hamming space. Although $\tanh(\cdot)$ encourages binarization, feature magnitudes often shrink, resulting in $\tanh(v) \approx v$ (see \cref{fig:tanh}). To enforce stronger binarization, we apply $l^2$-normalization and re-scale the features as $\tanh \left( s_1 \frac{v}{||v||_2} \right)$. The normalized Hamming distance (NHD) between the $i$-th sample and the memory feature $W_j$ is then defined as:
\begin{equation}
\label{eq:hamming_distance}
    {{d}_{ij}}=\frac{{{\left\| \tanh \left( {{s}_{1}}\frac{{{v}_{i}}}{{{\left\| {{v}_{i}} \right\|}_{2}}} \right)-\tanh \left( {{s}_{1}}\frac{{{W}_{j}}}{{{\left\| {{W}_{j}} \right\|}_{2}}} \right) \right\|}_{1}}}{l}.
\end{equation}
We then propose the loss as:
\begin{equation}
\label{eq:loss_HD}
    \LossNHD=-\log \frac{e^{s \frac{\left(2-d_{ii}\right)^2}{4} }}{ e^{s \frac{\left(2-d_{ii}\right)^2}{4}}+\sum\limits_{j=1,j\ne i}^{N}{e^{s{\left( 1-d_{ij} \right)}^2 }} }.
\end{equation}
The NHD between two random hashes follows a truncated normal distribution centered at 1 within $[0,2]$, as shown in \cref{fig:hamming}. Accordingly, $\left(1-d_{ij}\right)^2$ encourages negative pairs to behave like random vectors, while positive pairs maximize $\left(2-d_{ii}\right)^2$ by enforcing feature similarity. The term $\frac{(2-d_{ii})^2}{4}$ rescales the positive score to $[0,1]$, ensuring balanced contributions from positive and negative terms. By directly optimizing in the Hamming space, \LossNHD{} minimizes intra-instance distance while pushing different instances toward the random-hash regime, which theoretically reduces collision probability, as proven in the supplementary material.

While \LossNHD{} improves retrieval and reduces collisions by pushing samples apart, but may also separate semantically similar instances, requiring a debiasing step. To fairly evaluate our single-pass design, we adopt simple pseudo-labeling~\cite{hu2017pseudo} instead of more complex debiasing methods~\cite{wang2022contrastive}. In the unsupervised setting, we assume that the true number of clusters is unknown; therefore, we employ affinity propagation~\cite{frey2007clustering}, which automatically determines the number of clusters. Each cluster is treated as a pseudo class, and we introduce a learnable cluster memory $\Wpseudo \in \mathbb{R}^{n_c \times l}$, where each row represents a cluster center. Replacing $W$ with $\Wpseudo$ in~\cref{eq:hamming_distance,eq:loss_HD} yields a pseudo-label loss $\LossPseudo$ that pulls similar samples together to preserve semantic consistency. Pseudo labels are refreshed every $N_{\text{PEpochs}}$ epochs. Combined with \LossNHD{}, this pull-push scheme progressively refines the feature space for fine-grained retrieval. 

\subsection{Collision-sensitive attention}
\label{sec:attention}

Fine-grained image retrieval requires capturing subtle, localized visual differences often overlooked by global representations. We extract local features from the final convolutional layer or transformer patch tokens, forming a feature map $T \in \mathbb{R}^{C \times H \times W}$, viewed as $H \times W$ spatial features of dimension $C$. To emphasize cues critical for fine-grained discrimination and collision reduction, we introduce a Collision-Sensitive Attention (CSA) module that focuses on rare local patterns.

To distinguish common and rare local features, we introduce a learnable prototype vector $W^{(\text{att})} \in \mathbb{R}^{C}$ that represents globally frequent patterns across the dataset and is optimized to minimize its distance to spatial features as:
\begin{equation}
    \LossAttention = \sum_{i=1}^{H}\sum_{j=1}^{W} \left\| T_{ij} - W^{(\text{att})} \right\|_{1},
\end{equation}
where $T_{ij} \in \mathbb{R}^C$ is the feature at spatial location $(i,j)$. The learnable vector $W^{(\text{att})}$ represents common visual patterns, while the distance $||T_{ij} - W^{(\text{att})}||_1$ measures feature rarity, with larger distances indicating more distinctive local cues.

We then assign attention weights to each spatial position based on this rarity. Specifically, we compute the attention map $\alpha \in [0,1]^{H \times W}$ as the softmax-normalized rarity score:
\begin{equation}
    \alpha_{ij} = \frac{\exp(\left\| T_{ij} - W^{(\text{att})} \right\|_{1})}{\sum_{p=1}^{H}\sum_{q=1}^{W} \exp(\left\| T_{pq} - W^{(\text{att})} \right\|_{1})}.
\end{equation}
Features that deviate from the common prototype receive higher attention, encouraging the network to emphasize rare and collision-resistant cues. The attention-weighted features are aggregated by average pooling and concatenated with the global representation to form the final feature, which is fed into a fully connected layer to produce the $l$-bit hash code. This combined representation captures both global context and fine-grained local cues, enabling semantic consistency while effectively reducing hash collisions.

Overall, the proposed \ac{name} method is end-to-end trainable by combining all the proposed loss functions as follows:
\begin{equation}
\label{eq:loss_final}
    L= \LossNHD+ \LambdaPseudo \LossPseudo+ \LambdaAttention \LossAttention.
\end{equation}
Finally, at inference time, the final hash code $u$ is directly obtained from the feature vector $v$ via:
\begin{equation}
    u = \operatorname{sgn} \left( v \right).
\end{equation}

\section{Experiments}
\label{sec:experiments}

\begin{table*}[!htbp]
\caption{\small Retrieval \acs{mAP} (\%) comparison under the unsupervised learning scenario. The number below every dataset is the retrieval performance of continous features of pretrained VGG-16 and ViT-L/16 with dimensions 4096 and 1024, respectively using cosine similarity.}
\label{tab:main_results}
\centering
\resizebox{0.93\textwidth}{!}{
\begin{tabular}{cc|ccccccccc|ccc|ccc}
\hline
\multicolumn{2}{c|}{Backbone} & \multicolumn{9}{c|}{VGG16} & \multicolumn{3}{c|}{ResNet50} & \multicolumn{3}{c}{ViT-L/16} \\ \hline
\multicolumn{1}{c|}{Dataset} & \# bits & UGH & MLS3RUDH & Bihalf & CIBHash & SPQ & MeCoQ & SDC & $A^2$-SSL & \textbf{Our \ac{name}} & SPQ & SDC & \textbf{Our \ac{name}} & SPQ & SDC & \textbf{Our \ac{name}} \\ \hline
\multicolumn{1}{c|}{\multirow{5}{*}{\begin{tabular}[c]{@{}c@{}}Oxford Flowers\\ (34.60)\\ (92.64)\end{tabular}}} & 12 & 7.28 & 11.56 & 13.56 & 18.47 & 20.25 & 19.12 & 17.91 & 34.21 & \textbf{35.67} & 21.03 & 18.50 & \textbf{36.46} & 28.42 & 34.31 & \textbf{75.89} \\
\multicolumn{1}{c|}{} & 24 & 10.71 & 14.69 & 17.87 & 26.41 & 27.77 & 26.58 & 23.13 & 43.02 & \textbf{45.92} & 29.84 & 24.67 & \textbf{47.17} & 37.42 & 42.81 & \textbf{93.25} \\
\multicolumn{1}{c|}{} & 32 & 12.09 & 18.94 & 19.15 & 29.78 & 30.09 & 30.31 & 26.07 & 44.82 & \textbf{50.77} & 31.91 & 27.39 & \textbf{54.94} & 40.58 & 44.67 & \textbf{94.83} \\
\multicolumn{1}{c|}{} & 48 & 12.38 & 20.35 & 21.15 & 33.04 & 34.89 & 34.44 & 28.41 & 46.37 & \textbf{57.80} & 36.27 & 29.94 & \textbf{63.51} & 44.10 & 47.03 & \textbf{96.25} \\
\multicolumn{1}{c|}{} & 96 & 13.50 & 22.48 & 23.80 & 36.80 & 40.06 & 39.58 & 35.30 & - & \textbf{62.06} & 41.68 & 36.92 & \textbf{71.25} & 47.31 & 50.24 & \textbf{97.85} \\ \hline
\multicolumn{1}{c|}{\multirow{5}{*}{\begin{tabular}[c]{@{}c@{}}CUB200-2011\\ (21.84)\\ (63.05)\end{tabular}}} & 12 & 5.65 & 6.32 & 7.61 & 9.83 & 10.16 & 10.61 & 10.01 & 20.14 & \textbf{20.92} & 9.02 & 9.52 & \textbf{19.05} & 20.46 & 19.53 & \textbf{50.61} \\
\multicolumn{1}{c|}{} & 24 & 8.46 & 8.79 & 10.15 & 12.58 & 13.53 & 13.34 & 14.44 & 26.67 & \textbf{28.36} & 12.21 & 13.98 & \textbf{27.80} & 24.70 & 24.90 & \textbf{63.52} \\
\multicolumn{1}{c|}{} & 32 & 9.40 & 10.22 & 10.90 & 14.87 & 15.26 & 15.41 & 15.80 & 27.77 & \textbf{30.20} & 14.46 & 15.86 & \textbf{30.54} & 27.20 & 28.84 & \textbf{66.68} \\
\multicolumn{1}{c|}{} & 48 & 10.41 & 10.95 & 11.35 & 17.05 & 18.01 & 18.80 & 18.13 & 29.19 & \textbf{35.68} & 17.59 & 17.77 & \textbf{34.56} & 29.92 & 31.11 & \textbf{67.78} \\
\multicolumn{1}{c|}{} & 96 & 11.70 & 12.11 & 12.52 & 18.59 & 19.52 & 19.39 & 20.09 & - & \textbf{39.85} & 19.80 & 19.40 & \textbf{38.71} & 32.83 & 33.78 & \textbf{70.33} \\ \hline
\multicolumn{1}{c|}{\multirow{5}{*}{\begin{tabular}[c]{@{}c@{}}Stanford Cars\\ (5.60)\\ (8.81)\end{tabular}}} & 12 & 1.86 & 2.01 & 1.96 & 2.56 & 2.92 & 2.71 & 2.73 & 3.30 & \textbf{3.46} & 2.69 & 3.05 & \textbf{3.58} & 3.72 & 3.76 & \textbf{4.82} \\
\multicolumn{1}{c|}{} & 24 & 2.32 & 2.42 & 2.34 & 3.36 & 3.53 & 3.33 & 3.27 & 4.38 & \textbf{5.88} & 3.72 & 3.39 & \textbf{6.46} & 4.43 & 5.22 & \textbf{6.73} \\
\multicolumn{1}{c|}{} & 32 & 2.53 & 2.50 & 2.64 & 3.39 & 4.08 & 3.83 & 3.68 & 4.81 & \textbf{6.82} & 4.22 & 3.91 & \textbf{6.92} & 6.16 & 6.28 & \textbf{7.53} \\
\multicolumn{1}{c|}{} & 48 & 2.66 & 2.68 & 2.84 & 3.73 & 4.36 & 4.18 & 4.00 & 5.28 & \textbf{7.63} & 4.41 & 4.09 & \textbf{8.44} & 6.92 & 7.13 & \textbf{9.35} \\
\multicolumn{1}{c|}{} & 96 & 2.89 & 2.91 & 3.09 & 4.06 & 4.67 & 4.49 & 4.29 & - & \textbf{8.32} & 5.08 & 4.95 & \textbf{9.35} & 7.72 & 7.91 & \textbf{11.07} \\ \hline
\multicolumn{1}{c|}{\multirow{5}{*}{\begin{tabular}[c]{@{}c@{}}Stanford Dogs\\ (58.56)\\ (81.41)\end{tabular}}} & 12 & 23.97 & 30.08 & 32.42 & 34.55 & 38.94 & 36.16 & 35.34 & 54.80 & \textbf{55.16} & 36.75 & 36.21 & \textbf{54.80} & 38.21 & 39.42 & \textbf{63.25} \\
\multicolumn{1}{c|}{} & 24 & 35.35 & 40.17 & 44.34 & 46.14 & 48.52 & 46.96 & 47.96 & 61.20 & \textbf{64.05} & 50.87 & 49.28 & \textbf{64.10} & 50.91 & 51.50 & \textbf{82.15} \\
\multicolumn{1}{c|}{} & 32 & 39.55 & 44.00 & 48.57 & 48.96 & 53.21 & 50.45 & 52.13 & 62.70 & \textbf{65.74} & 53.76 & 53.99 & \textbf{65.42} & 54.12 & 54.73 & \textbf{84.66} \\
\multicolumn{1}{c|}{} & 48 & 44.10 & 48.22 & 51.88 & 51.67 & 56.55 & 53.76 & 55.56 & 64.56 & \textbf{69.22} & 57.92 & 56.44 & \textbf{69.39} & 57.87 & 58.36 & \textbf{88.33} \\
\multicolumn{1}{c|}{} & 96 & 47.98 & 52.07 & 55.17 & 54.45 & 60.31 & 56.96 & 59.27 & - & \textbf{71.70} & 61.34 & 60.77 & \textbf{70.30} & 61.15 & 61.92 & \textbf{89.12} \\ \hline
\multicolumn{1}{c|}{\multirow{5}{*}{\begin{tabular}[c]{@{}c@{}}Food101\\ (10.80)\\ (42.44)\end{tabular}}} & 12 & 4.32 & 5.11 & 6.20 & 7.27 & 7.95 & 7.87 & 7.51 & 8.48 & \textbf{10.54} & 8.12 & 7.89 & \textbf{11.55} & 12.91 & 12.52 & \textbf{35.16} \\
\multicolumn{1}{c|}{} & 24 & 6.33 & 7.45 & 8.25 & 10.62 & 11.22 & 10.95 & 9.50 & 12.89 & \textbf{16.26} & 12.34 & 10.92 & \textbf{19.72} & 14.18 & 14.77 & \textbf{48.12} \\
\multicolumn{1}{c|}{} & 32 & 7.29 & 8.63 & 9.24 & 11.74 & 12.20 & 11.56 & 10.43 & 14.08 & \textbf{18.45} & 13.94 & 11.62 & \textbf{22.23} & 15.94 & 16.38 & \textbf{52.14} \\
\multicolumn{1}{c|}{} & 48 & 8.40 & 9.59 & 10.18 & 12.17 & 13.33 & 13.24 & 11.64 & 15.38 & \textbf{20.16} & 15.63 & 13.21 & \textbf{25.36} & 17.77 & 18.22 & \textbf{54.45} \\
\multicolumn{1}{c|}{} & 96 & 9.39 & 10.59 & 11.29 & 13.41 & 14.90 & 14.78 & 12.94 & - & \textbf{23.45} & 15.97 & 14.52 & \textbf{26.93} & 21.90 & 22.41 & \textbf{57.06} \\ \hline
\multicolumn{1}{c|}{\multirow{3}{*}{\begin{tabular}[c]{@{}c@{}}NUS-WIDE\\ (82.53)\\ (82.51)\end{tabular}}} & 16 & 63.3 & 71.3 & 77.4 & 77.1 & 76.8 & 80.2 & 80.7 & - & \textbf{80.9} & 76.6 & 81.2 & \textbf{81.3} & 80.2 & 81.1 & \textbf{81.4} \\
\multicolumn{1}{c|}{} & 32 & 69.1 & 72.7 & 80.1 & 79.7 & 78.6 & 82.2 & 82.3 & - & \textbf{83.7} & 77.4 & 83.2 & \textbf{83.7} & 83.0 & 83.4 & \textbf{83.9} \\
\multicolumn{1}{c|}{} & 64 & 73.1 & 75.0 & 81.9 & 80.9 & 81.3 & 83.2 & 83.4 & - & \textbf{84.9} & 78.5 & 84.2 & \textbf{85.0} & 83.9 & 84.0 & \textbf{84.8} \\ \hline
\end{tabular}
}
\end{table*}

This section evaluates the proposed method under a unified setup. Following standard protocols~\cite{qiu2021unsupervised, wang2022contrastive, hu2024asymmetric}, the images are resized to $224\times224$ and trained with common data augmentations. The models are trained for $100$ epochs with scale parameters set to $s=s_1=8$ and pseudo labels refreshed every $5$ epochs. We evaluate three backbones, VGG-16~\cite{simonyan2014very}, ResNet-50~\cite{he2016deep}, and ViT-L/16~\cite{dosovitskiy2020image}. We conduct experiments on five fine-grained benchmarks (CUB200-2011~\cite{wah2011caltech}, Oxford Flowers~\cite{nilsback2008automated}, Stanford Dogs~\cite{khosla2011novel}, Stanford Cars \cite{krause20133d}, and Food101~\cite{bossard2014food}) and one wide multi-label dataset (NUS-WIDE~\cite{chua2009nus}). Additional details are provided in the appendix.

\subsection{Unsupervised learning performance}
\label{sec:un_performance}

In this experiment, we evaluate the proposed method against several strong baselines, including UGH \cite{li2021deep}, MLS3RUDH \cite{tu2020mls3rduh}, CIBHash \cite{qiu2021unsupervised}, SPQ \cite{jang2021self}, MeCoQ \cite{wang2022contrastive}, SDC \cite{ng2023unsupervised}, and $A^2$-SSL \cite{hu2024asymmetric}. We compare only with unsupervised methods, with $A^2$-SSL being the most recent, whose results are reported as provided since no public source code is available. Most baseline results are taken from \cite{hu2024asymmetric} or we used their official implementations. \Cref{tab:main_results} reports the mAP (\%) across datasets, evaluated at 12, 24, 32, 48, and 96 bits for fine-grained datasets, and 16, 32, and 64 bits for the wide dataset, following prior works~\cite{jang2021self, wang2022contrastive, ng2023unsupervised}. Our \ac{name} consistently outperforms all baselines across datasets and code lengths.

\subsection{Hard-unsupervised learning performance}

\begin{table}[!ht]
\centering
\caption{\small Retrieval \acs{mAP} (\%) comparison under the hard-unsupervised setting. "RN", "VGG", and "ViT" denote ResNet50, VGG16, and ViT-L/16 backbones, respectively.}
\label{tab:hard_reuslts}
\resizebox{0.88\linewidth}{!}{
\begin{tabular}{c|c|ccc|cc}
\hline
Dataset & \# bits & SGH & SPQ-RN & Our-RN & Our-VGG & Our-ViT \\ \hline
\multirow{5}{*}{Oxford Flowers} & 12 & 3.52 & 12.56 & 23.85 & 22.74 & \textbf{24.91} \\
 & 24 & 4.38 & 24.74 & \textbf{30.91} & 29.12 & 28.57 \\
 & 32 & 8.16 & 27.89 & \textbf{36.25} & 34.36 & 33.21 \\
 & 48 & 10.42 & 31.78 & \textbf{40.71} & 42.83 & 41.42 \\
 & 96 & 12.71 & 36.02 & 44.82 & \textbf{46.58} & 45.33 \\ \hline
\multirow{5}{*}{CUB200-2011} & 12 & 2.09 & 6.52 & \textbf{15.62} & 14.87 & 14.23 \\
 & 24 & 2.82 & 10.89 & \textbf{22.63} & 21.32 & 20.74 \\
 & 32 & 2.94 & 13.74 & \textbf{26.48} & 25.11 & 24.56 \\
 & 48 & 5.65 & 15.63 & \textbf{29.73} & 27.94 & 27.22 \\
 & 96 & 8.05 & 17.09 & \textbf{33.92} & 31.87 & 31.15 \\ \hline
\multirow{5}{*}{Stanford Cars} & 12 & 0.94 & 1.86 & \textbf{2.94} & 2.73 & 2.65 \\
 & 24 & 1.57 & 3.02 & \textbf{5.41} & 5.05 & 4.88 \\
 & 32 & 2.01 & 3.57 & \textbf{6.33} & 5.91 & 5.64 \\
 & 48 & 2.73 & 4.32 & \textbf{7.11} & 6.75 & 6.46 \\
 & 96 & 3.22 & 5.08 & \textbf{7.86} & 7.42 & 7.21 \\ \hline
\multirow{5}{*}{Stanford Dogs} & 12 & 17.42 & 31.58 & \textbf{36.52} & 34.63 & 33.91 \\
 & 24 & 24.28 & 42.96 & 48.90 & \textbf{49.32} & 46.10 \\
 & 32 & 26.11 & 47.44 & \textbf{53.54} & 52.25 & 51.84 \\
 & 48 & 28.30 & 51.79 & 57.82 & \textbf{58.41} & 55.93 \\
 & 96 & 31.86 & 52.13 & 60.21 & \textbf{62.64} & 58.02 \\ \hline
\multirow{5}{*}{Food101} & 12 & 3.81 & 7.03 & \textbf{9.84} & 9.25 & 9.41 \\
 & 24 & 4.83 & 10.81 & 13.12 & \textbf{13.86} & 14.07 \\
 & 32 & 7.05 & 9.35 & 14.48 & 15.03 & \textbf{15.21} \\
 & 48 & 8.26 & 10.98 & 15.63 & 15.32 & \textbf{15.64} \\
 & 96 & 9.12 & 12.46 & 16.84 & 16.74 & \textbf{17.02} \\ \hline
\multirow{3}{*}{NUS-WIDE} & 16 & 59.3 & 76.6 & \textbf{78.45} & 78.12 & 78.00 \\
 & 32 & 59.0 & 77.4 & \textbf{79.22} & 78.90 & 78.63 \\
 & 64 & 60.7 & 78.5 & 79.86 & \textbf{80.43} & 80.12 \\ \hline
\end{tabular}
}
\end{table}

Hard-unsupervised learning refers to training models entirely from scratch without pretrained backbones~\cite{jang2021self}. Since only few methods explicitly follow this setting and $A^2$-SSL does not provide public source code, we compare our approach with SGH~\cite{dai2017stochastic} and SPQ~\cite{jang2021self} on the same datasets as in~\cref{sec:un_performance}. \Cref{tab:hard_reuslts} reports the \ac{mAP} (\%) under the hard-unsupervised setting. \ac{name} consistently outperforms all baselines, demonstrating its ability to learn discriminative fine-grained representations from scratch without pretrained backbones.

\subsection{Collision}
In this experiment, we evaluate the collision resistance of \ac{name} compared to several methods with VGG16 backbone. We select three datasets with different scales and characteristics: CUB200-2011, Stanford Dogs, and NUS-WIDE.

\begin{table}[htbp]
\caption{\small Collision probability $\Pcollision$ (\textpertenthousand) comparison. "*" denotes the hard-unsupervised setting.}
\label{tab:collision_results}
\centering
\resizebox{0.95\linewidth}{!}{
\begin{tabular}{cc|c|cccccc}
\hline
\multicolumn{2}{c|}{Dataset}                                                     & \# bits & SPQ   & MeCoQ & SDC   & SPQ*  & Our \acs{name} & Our \acs{name}* \\ \hline
\multicolumn{1}{c|}{\multirow{10}{*}{CUB200-2011}}   & \multirow{5}{*}{Database} & 12      & 198.7 & 184.6 & 176.4 & 130.5 & {\ul 9.42}     & \textbf{6.85}   \\
\multicolumn{1}{c|}{}                                &                           & 24      & 182.5 & 174.1 & 169.8 & 123.9 & {\ul 3.59}     & \textbf{2.40}   \\
\multicolumn{1}{c|}{}                                &                           & 32      & 130.1 & 162.2 & 142.6 & 112.2 & \textbf{1.12}  & {\ul 1.45}      \\
\multicolumn{1}{c|}{}                                &                           & 48      & 38.5  & 8.0   & 9.3   & 12.6  & \textbf{0.27}  & {\ul 0.35}      \\
\multicolumn{1}{c|}{}                                &                           & 96      & 8.4   & 8.1   & 6.7   & 7.4   & \textbf{0.02}  & {\ul 0.03}      \\ \cline{2-9} 
\multicolumn{1}{c|}{}                                & \multirow{5}{*}{Testset}  & 12      & 205.3 & 191.2 & 180.6 & 135.4 & \textbf{8.74}  & {\ul 10.23}     \\
\multicolumn{1}{c|}{}                                &                           & 24      & 186.3 & 180.4 & 174.4 & 134.7 & {\ul 2.98}     & \textbf{2.09}   \\
\multicolumn{1}{c|}{}                                &                           & 32      & 135.9 & 171.0 & 143.0 & 116.8 & \textbf{0.68}  & {\ul 0.81}      \\
\multicolumn{1}{c|}{}                                &                           & 48      & 40.0  & 8.1   & 10.9  & 12.4  & {\ul 0.22}     & \textbf{0.19}   \\
\multicolumn{1}{c|}{}                                &                           & 96      & 8.6   & 8.9   & 7.4   & 7.5   & \textbf{0.02}  & {\ul 0.05}      \\ \hline
\multicolumn{1}{c|}{\multirow{10}{*}{Stanford Dogs}} & \multirow{5}{*}{Database} & 12      & 112.5 & 127.8 & 118.4 & 90.6  & \textbf{11.21} & {\ul 11.30}     \\
\multicolumn{1}{c|}{}                                &                           & 24      & 45.5  & 53.3  & 47.4  & 73.4  & {\ul 3.83}     & \textbf{3.34}   \\
\multicolumn{1}{c|}{}                                &                           & 32      & 23.5  & 37.5  & 38.6  & 27.6  & \textbf{2.16}  & {\ul 2.31}      \\
\multicolumn{1}{c|}{}                                &                           & 48      & 18.5  & 21.2  & 23.6  & 14.5  & \textbf{0.43}  & {\ul 0.46}      \\
\multicolumn{1}{c|}{}                                &                           & 96      & 4.8   & 4.4   & 5.1   & 4.5   & \textbf{0.20}  & \textbf{0.20}   \\ \cline{2-9} 
\multicolumn{1}{c|}{}                                & \multirow{5}{*}{Testset}  & 12      & 120.9 & 135.2 & 124.5 & 95.8  & {\ul 10.51}    & \textbf{9.99}   \\
\multicolumn{1}{c|}{}                                &                           & 24      & 46.9  & 55.5  & 48.2  & 75.9  & {\ul 3.97}     & \textbf{3.60}   \\
\multicolumn{1}{c|}{}                                &                           & 32      & 26.7  & 39.2  & 40.9  & 22.6  & {\ul 1.92}     & \textbf{1.88}   \\
\multicolumn{1}{c|}{}                                &                           & 48      & 20.8  & 24.8  & 23.5  & 11.9  & \textbf{0.51}  & {\ul 0.56}      \\
\multicolumn{1}{c|}{}                                &                           & 96      & 4.8   & 4.5   & 5.5   & 4.7   & \textbf{0.14}  & {\ul 0.15}      \\ \hline
\multicolumn{1}{c|}{\multirow{6}{*}{NUS-WIDE}}       & \multirow{3}{*}{Database} & 16      & 199.4 & 157.7 & 122.2 & 115.9 & {\ul 5.46}     & \textbf{4.79}   \\
\multicolumn{1}{c|}{}                                &                           & 32      & 245.3 & 121.2 & 175.9 & 177.9 & {\ul 1.16}     & \textbf{0.73}   \\
\multicolumn{1}{c|}{}                                &                           & 64      & 17.2  & 38.8  & 20.0  & 15.3  & {\ul 0.44}     & \textbf{0.34}   \\ \cline{2-9} 
\multicolumn{1}{c|}{}                                & \multirow{3}{*}{Testset}  & 16      & 213.3 & 174.4 & 144.4 & 127.9 & {\ul 11.30}    & \textbf{9.81}   \\
\multicolumn{1}{c|}{}                                &                           & 32      & 265.9 & 130.8 & 179.1 & 204.4 & {\ul 1.28}     & \textbf{0.71}   \\
\multicolumn{1}{c|}{}                                &                           & 64      & 19.8  & 33.8  & 24.0  & 16.9  & {\ul 0.47}     & \textbf{0.25}   \\ \hline
\end{tabular}
}
\end{table}

\Cref{tab:collision_results} summarizes the $\Pcollision$ at different hash code lengths. The lower the collision probability, the higher the collision resistance. Our \ac{name} consistently achieves the lowest collision rates across all settings. Notably, both the best and second-best results are achieved by our method under different training scenarios, demonstrating its consistent effectiveness in preserving diverse binary codes.

\subsection{Complexity}

We evaluate complexity under four settings: medium-scale (CUB200-2011), large-scale (Food101), and stress tests at $N=1$M and $N=10$M images, as summarized in~\cref{tab:complexity_results}. Stress tests are included only for scalability analysis, as no real fine-grained datasets exist at this scale. Our method consistently achieves the lowest complexity across all settings and backbones, with computation dominated by the backbone and negligible overhead from the memory bank. Although its complexity grows slightly with $N$, this increase is minimal and practically insignificant. In practice, existing unsupervised hashing methods rarely train on full large-scale datasets. At inference, our method directly applies $\operatorname{sgn}(\cdot)$, resulting in lowest complexity compared to clustering-based methods~\cite{jang2021self, wang2022contrastive} (with $2^{l/M}\!\cdot\!M$ centroids), yielding the overall lowest complexity.

\begin{table}[!htbp]
\caption{\small Complexity across dataset sizes ($N$ images). Numbers in parentheses indicate the complexity for the backbone itself.}
\label{tab:complexity_results}
\centering
\resizebox{0.95\linewidth}{!}{
\begin{tabular}{l|cc|cc}
\hline
\multirow{2}{*}{Method} & \multicolumn{2}{c|}{Time complexity (GFLOPs)} & \multicolumn{2}{c}{Space complexity (GB)} \\ \cline{2-5} 
                        & VGG16 (15.47)        & ViT-L/16 (61.55)       & VGG16 (8.48)      & ViT-L/16 (19.93)      \\ \hline
SPQ                     & 30.95                & 123.1                  & 16.96             & 39.86                 \\
MeCoQ                   & 30.95                & 123.1                  & 16.96             & 39.86                 \\
$A^2$-SSL               & 46.44                & 184.7                  & 25.54             & 59.89                 \\ \hline
\acs{name} (N=5,994)    & 15.47                & 61.55                  & 8.48              & 19.93                 \\
\acs{name} (N=75,750)   & 15.47                & 61.55                  & 8.48              & 19.93                 \\
\acs{name} (N=1M)       & 15.47                & 61.55                  & 8.61              & 20.06                 \\
\acs{name} (N=10M)      & 15.79                & 61.87                  & 9.76              & 21.21                 \\ \hline
\end{tabular}
}
\end{table}

\subsection{Ablation Studies}
\begin{table}[ht]
\caption{\small Retrieval \ac{mAP} (\%) of different configurations of \ac{name}.}
\label{tab:ablation_retrieval}
\centering
\resizebox{0.95\linewidth}{!}{
\begin{tabular}{lcccccc}
\hline
\multicolumn{1}{l|}{\multirow{2}{*}{Setting}} & \multicolumn{3}{c|}{CUB200-2011} & \multicolumn{3}{c}{Oxford Flowers} \\
\multicolumn{1}{l|}{} & 12-bit & 32-bit & \multicolumn{1}{c|}{96-bit} & 12-bit & 32-bit & 96-bit \\ \hline
\multicolumn{7}{c}{Unsupervised} \\ \hline
\multicolumn{1}{l|}{Ours w/o pseudo, CSA} & 2.93 & 8.43 & \multicolumn{1}{c|}{18.67} & 6.54 & 33.01 & 55.57 \\
\multicolumn{1}{l|}{Ours w/o CSA} & 12.08 & 23.37 & \multicolumn{1}{c|}{31.05} & 25.05 & 46.25 & 59.20 \\
\multicolumn{1}{l|}{Ours w/o global} & 17.08 & 26.83 & \multicolumn{1}{c|}{38.16} & 33.10 & 47.76 & 61.21 \\
\multicolumn{1}{l|}{Ours $\lambda_{\text{att}}=0.5$} & 18.46 & 28.94 & \multicolumn{1}{c|}{38.51} & 33.28 & 49.83 & 61.74 \\
\multicolumn{1}{l|}{Ours $\lambda_{\text{pseudo}}=0.5$} & 19.73 & 29.87 & \multicolumn{1}{c|}{39.02} & 34.82 & 48.02 & 60.18 \\
\multicolumn{1}{l|}{Ours $\lambda_{\text{att}}=2.0$} & 19.18 & 28.62 & \multicolumn{1}{c|}{38.24} & 33.94 & 49.32 & 61.11 \\
\multicolumn{1}{l|}{Ours $\lambda_{\text{pseudo}}=2.0$} & \textbf{21.50} & 29.45 & \multicolumn{1}{c|}{38.77} & \textbf{37.20} & 49.91 & 61.56 \\
\multicolumn{1}{l|}{Ours $N_{\text{PEpochs}}=10$} & 19.12 & 30.06 & \multicolumn{1}{c|}{38.54} & 35.20 & {\ul 50.65} & \textbf{64.16} \\
\multicolumn{1}{l|}{Ours $s=16$, $s_1=8$} & 18.61 & 29.81 & \multicolumn{1}{c|}{38.12} & 34.97 & 50.31 & 61.48 \\
\multicolumn{1}{l|}{Ours $s=8$, $s_1=16$} & 18.35 & \textbf{31.96} & \multicolumn{1}{c|}{{\ul 39.07}} & {\ul 35.82} & 50.55 & {\ul 62.30} \\
\multicolumn{1}{l|}{Ours \acs{name}} & {\ul 20.92} & {\ul 30.20} & \multicolumn{1}{c|}{\textbf{39.85}} & 35.67 & \textbf{50.77} & 62.06 \\ \hline
\multicolumn{7}{c}{Hard-unsupervised} \\ \hline
\multicolumn{1}{l|}{Ours w/o pseudo, CSA} & 2.14 & 6.15 & \multicolumn{1}{c|}{13.92} & 8.30 & 12.33 & 25.41 \\
\multicolumn{1}{l|}{Ours w/o CSA} & 10.24 & 18.46 & \multicolumn{1}{c|}{25.27} & 14.93 & 29.85 & 41.93 \\
\multicolumn{1}{l|}{Ours w/o global} & {\ul 12.28} & {\ul 24.97} & \multicolumn{1}{c|}{\textbf{34.56}} & {\ul 20.71} & \textbf{37.83} & {\ul 44.14} \\
\multicolumn{1}{l|}{Ours \acs{name}} & \textbf{14.87} & \textbf{25.11} & \multicolumn{1}{c|}{{\ul 31.87}} & \textbf{22.74} & {\ul 34.36} & \textbf{46.58} \\ \hline
\end{tabular}
}
\end{table}

\begin{table}[ht]
\caption{\small Collision $\Pcollision$ (\textpertenthousand) of different settings of \ac{name}.}
\label{tab:ablation_collision}
\centering
\resizebox{0.95\linewidth}{!}{
\begin{tabular}{lcccccc}
\hline
\multicolumn{1}{l|}{\multirow{2}{*}{Setting}} & \multicolumn{3}{c|}{CUB200-2011} & \multicolumn{3}{c}{Oxford Flowers} \\
\multicolumn{1}{l|}{} & 12-bit & 32-bit & \multicolumn{1}{c|}{96-bit} & 12-bit & 32-bit & 96-bit \\ \hline
\multicolumn{7}{c}{Unsupervised} \\ \hline
\multicolumn{1}{l|}{Ours w/o pseudo, CSA} & \textbf{3.63} & \textbf{0.25} & \multicolumn{1}{c|}{\textbf{0.02}} & \textbf{4.68} & \textbf{0.13} & \textbf{0.01} \\
\multicolumn{1}{l|}{Ours w/o CSA} & 35.43 & 7.01 & \multicolumn{1}{c|}{0.80} & 41.94 & 8.38 & 0.84 \\
\multicolumn{1}{l|}{Ours w/o global} & {\ul 5.56} & {\ul 0.33} & \multicolumn{1}{c|}{{\ul 0.03}} & 13.35 & 0.58 & \textbf{0.01} \\
\multicolumn{1}{l|}{Ours $\LambdaAttention=0.5$} & 9.84 & 0.86 & \multicolumn{1}{c|}{0.04} & 14.33 & 0.79 & 0.03 \\
\multicolumn{1}{l|}{Ours $\LambdaPseudo=0.5$} & 7.96 & 0.52 & \multicolumn{1}{c|}{0.06} & 14.24 & {\ul 0.51} & 0.05 \\
\multicolumn{1}{l|}{Ours $\LambdaAttention=2.0$} & 9.31 & 0.83 & \multicolumn{1}{c|}{{\ul 0.03}} & 13.80 & 0.65 & 0.03 \\
\multicolumn{1}{l|}{Ours $\LambdaPseudo=2.0$} & 14.47 & 1.39 & \multicolumn{1}{c|}{0.08} & 19.82 & 1.24 & 0.06 \\
\multicolumn{1}{l|}{Ours $N_{\text{PEpoch}}=10$} & 16.02 & 1.75 & \multicolumn{1}{c|}{0.10} & 22.66 & 1.58 & 0.09 \\
\multicolumn{1}{l|}{Ours $s=16$, $s_1=8$} & 8.91 & 0.79 & \multicolumn{1}{c|}{{\ul 0.03}} & 12.95 & 0.70 & 0.03 \\
\multicolumn{1}{l|}{Ours $s=8$, $s_1=16$} & 8.74 & 0.72 & \multicolumn{1}{c|}{0.04} & {\ul 12.46} & 0.75 & {\ul 0.02} \\
\multicolumn{1}{l|}{Ours \acs{name}} & 8.74 & 0.68 & \multicolumn{1}{c|}{\textbf{0.02}} & 16.80 & 0.63 & {\ul 0.02} \\ \hline
\multicolumn{7}{c}{Hard-unsupervised} \\ \hline
\multicolumn{1}{l|}{Ours w/o pseudo, CSA} & {\ul 12.64} & 1.41 & \multicolumn{1}{c|}{\textbf{0.03}} & \textbf{5.83} & \textbf{0.08} & \textbf{0.0} \\
\multicolumn{1}{l|}{Ours w/o CSA} & 38.25 & 9.03 & \multicolumn{1}{c|}{1.02} & 37.34 & 9.26 & 0.72 \\
\multicolumn{1}{l|}{Ours w/o global} & 15.80 & {\ul 1.29} & \multicolumn{1}{c|}{\textbf{0.03}} & {\ul 10.75} & {\ul 0.32} & \textbf{0.0} \\
\multicolumn{1}{l|}{Ours \acs{name}} & \textbf{10.23} & \textbf{0.81} & \multicolumn{1}{c|}{{\ul 0.05}} & 17.87 & 0.84 & {\ul 0.08} \\ \hline
\end{tabular}
}
\end{table}

We conduct ablation studies on two widely used fine-grained datasets~\cite{hu2024asymmetric}, CUB200-2011 and Oxford Flowers, using the VGG-16 backbone. As shown in \cref{tab:ablation_retrieval,tab:ablation_collision}, each component contributes to overall performance. \LossNHD{} provides strong collision resistance but lower retrieval accuracy at short code lengths, where the number of possible binary codes ($2^{12}=4096$) is smaller than the dataset size. Additional analysis of the retrieval-collision trade-off is provided in the supplementary material (\cref{sec:setup}). While $L_{\text{pseudo}}$ improves retrieval accuracy but raises the collision rate. The CSA mitigates both issues, reducing collisions while boosting retrieval performance. Together, these components form a balanced and robust framework that remains stable across different settings.

\section{Conclusion}
\label{sec:conclusion}

This paper presented \ac{name}, a collision-resistant single-pass framework for unsupervised fine-grained image hashing. By eliminating the need for multi-view contrastive learning and directly optimizing in the Hamming space, our method reduces computational cost while maintaining high retrieval accuracy. We further adapt collision resistance from cryptography to deep hashing and introduce a collision-sensitive attention module that emphasizes rare discriminative regions. Extensive experiments show that \ac{name} achieves superior retrieval performance with the lowest collision rates and computational complexity among existing methods.

\clearpage
\setcounter{page}{1}

In the supplementary materials, we provide additional details and analyses to complement the main paper. We first present supporting materials for the proposed method in \cref{sec:details}. We then describe the experimental setup and dataset statistics in \cref{sec:setup}. Next, \cref{sec:codebook} presents an extension of the proposed single-pass framework to a clustering-based (codebook) formulation. In \cref{sec:proof}, we provide a formal mathematical proof analyzing the behavior of the normalized Hamming distance loss and its collision-resistant property. Qualitative retrieval results and attention visualizations across datasets and backbones are shown in \cref{sec:vis}. Finally, \cref{sec:discussion} discusses limitations of the proposed method and directions for future work.

\section{Method details}
\label{sec:details}

\begin{figure}[h]
    \centering
    \includegraphics[width=0.52\linewidth]{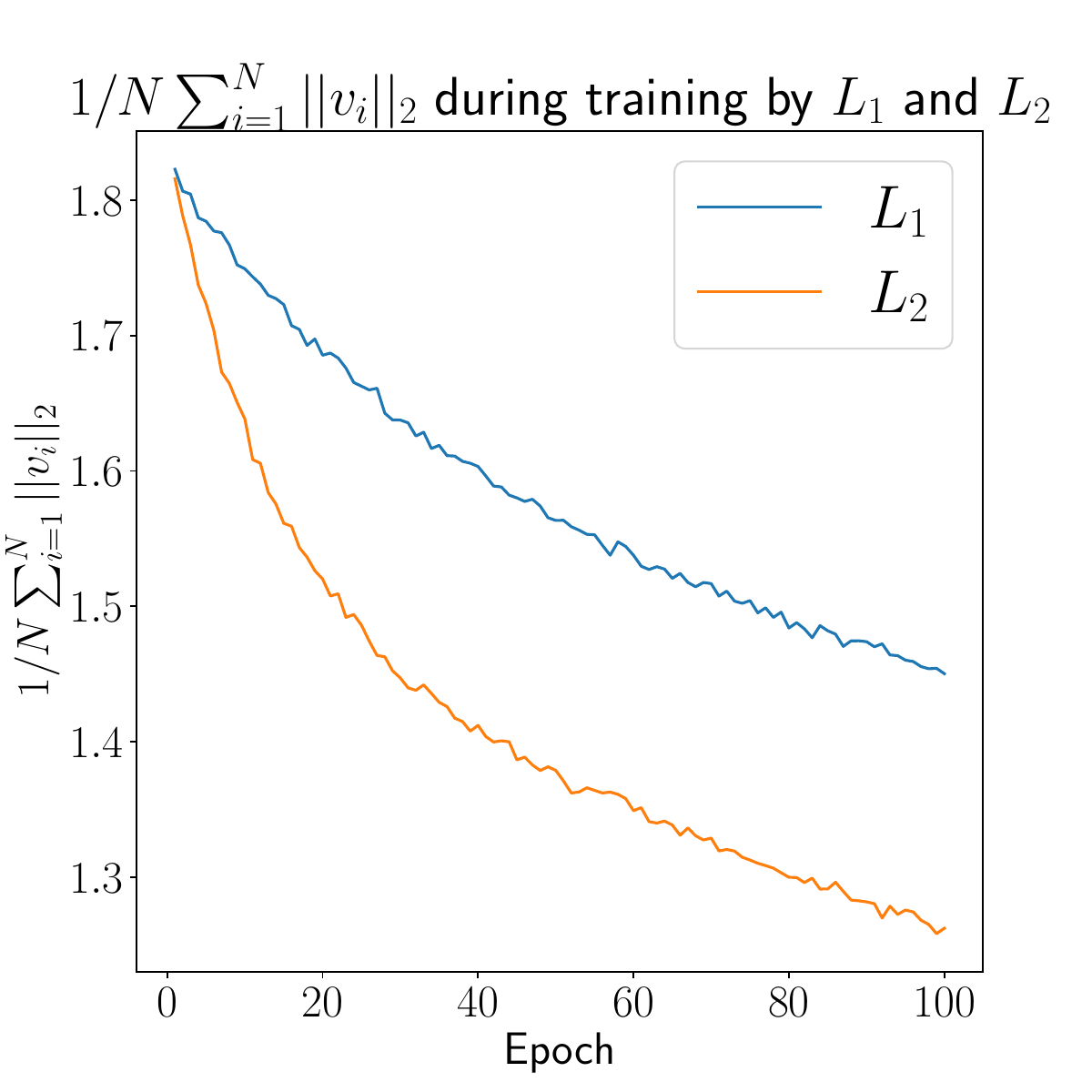}
    \caption{$\frac{1}{N}\sum_{i=1}^{N} ||v_i||_2$ during training by $L_1$ and $L_2$ on CUB200-2011, 100 epochs, $l=48$.}
    \label{fig:tanh}
\end{figure}

This section provides additional statistical evidence to support the design choices in \cref{eq:loss_HD}. \Cref{fig:tanh} illustrates the evolution of the average feature norm $\frac{1}{N}\sum_{i=1}^{N}\lVert v_i\rVert_2$ during training when using dot-product based objectives ($L_1$, $L_2$). We observe that the network tends to shrink feature magnitudes, causing $\tanh(v)\approx v$ and weakening the binarization effect. This behavior motivates the use of $\ell_2$ normalization before the $\tanh(\cdot)$ operation in our formulation. The scale parameter serves a standard re-scaling role and the key factor is normalization, which stabilizes feature magnitudes and enforces alignment with the Hamming space.

\begin{figure}[h]
    \centering
    \includegraphics[width=0.6\linewidth]{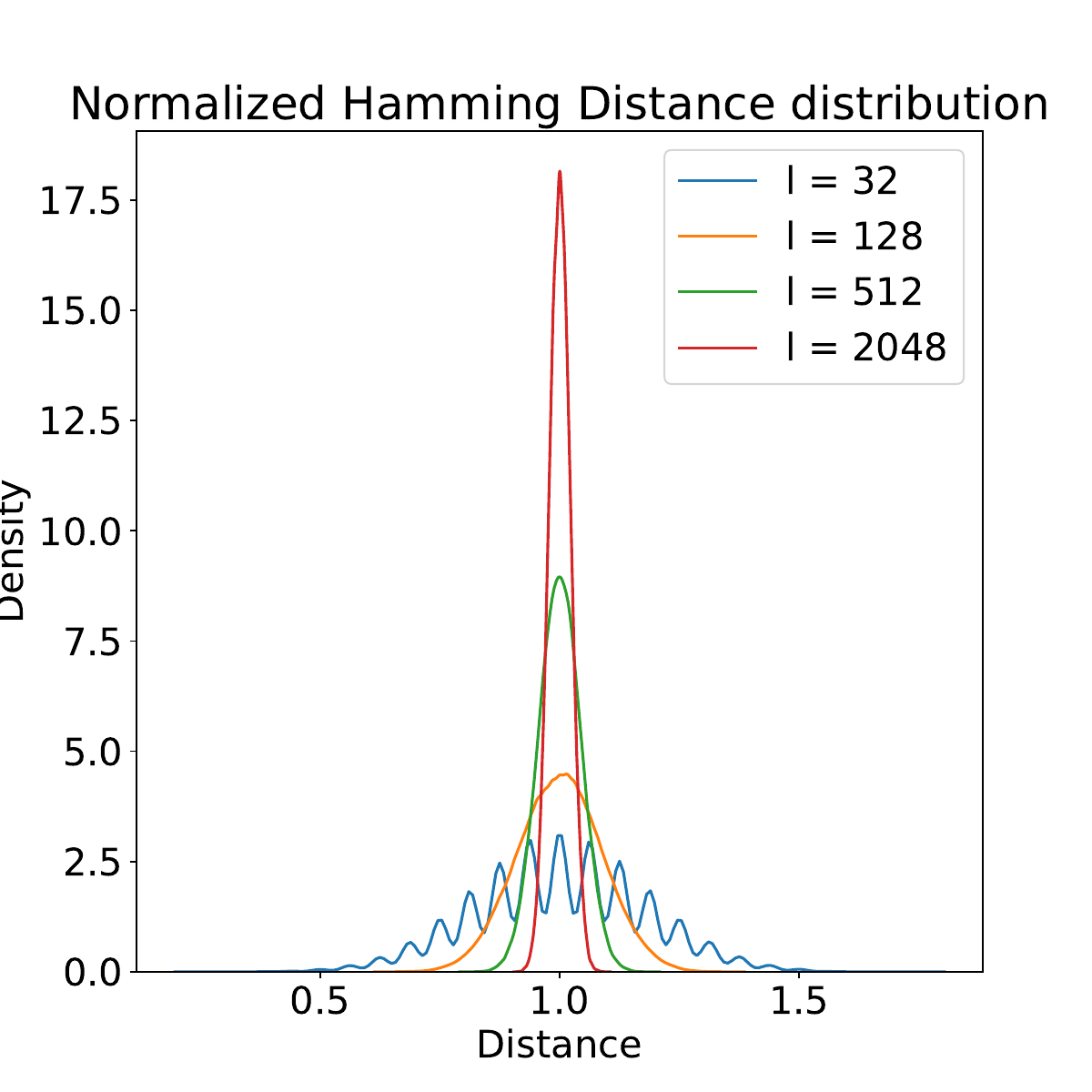}
    \caption{Distribution of $\text{NHD} \left( a, b \right)$ between two random vectors  $a,b \in \left\{-1, 1 \right\}^l$, $1000$ samples.}
    \label{fig:hamming}
\end{figure}

\Cref{fig:hamming} illustrates the empirical distribution of the normalized Hamming distance between pairs of random binary vectors for different code lengths. Regardless of $l$, the distance concentrates around $1$ and lies within the bounded range $[0,2]$, closely matching a truncated normal distribution. This behavior arises because each bit between two independent random binary vectors has equal probability of agreement and disagreement, making the expected normalized Hamming distance converge to $1$ as the code length increases. By the central limit theorem, the sum of independent bit differences tends toward a normal distribution. However, unlike an unconstrained Gaussian distribution, the normalized Hamming distance is strictly bounded within $[0,2]$, since every bit can contribute at most $2$ to the $\ell_1$ distance between vectors in $\{-1,1\}^l$. As a result, the Gaussian-like distribution is effectively truncated at both boundaries, producing the observed truncated normal behavior. This statistical property underpins the formulation of \LossNHD{}, where negative pairs are encouraged to behave like random hashes while positive pairs are pulled closer. Together, these analyses provide empirical justification for directly optimizing similarity in the Hamming space and for the specific normalization and re-scaling strategy adopted in our framework.

\begin{figure}[ht]
\centering
    \subfloat[Other multi-view method]{
        \begin{minipage}[b]{0.8\linewidth}
            \centering
            \includegraphics[width=\linewidth]{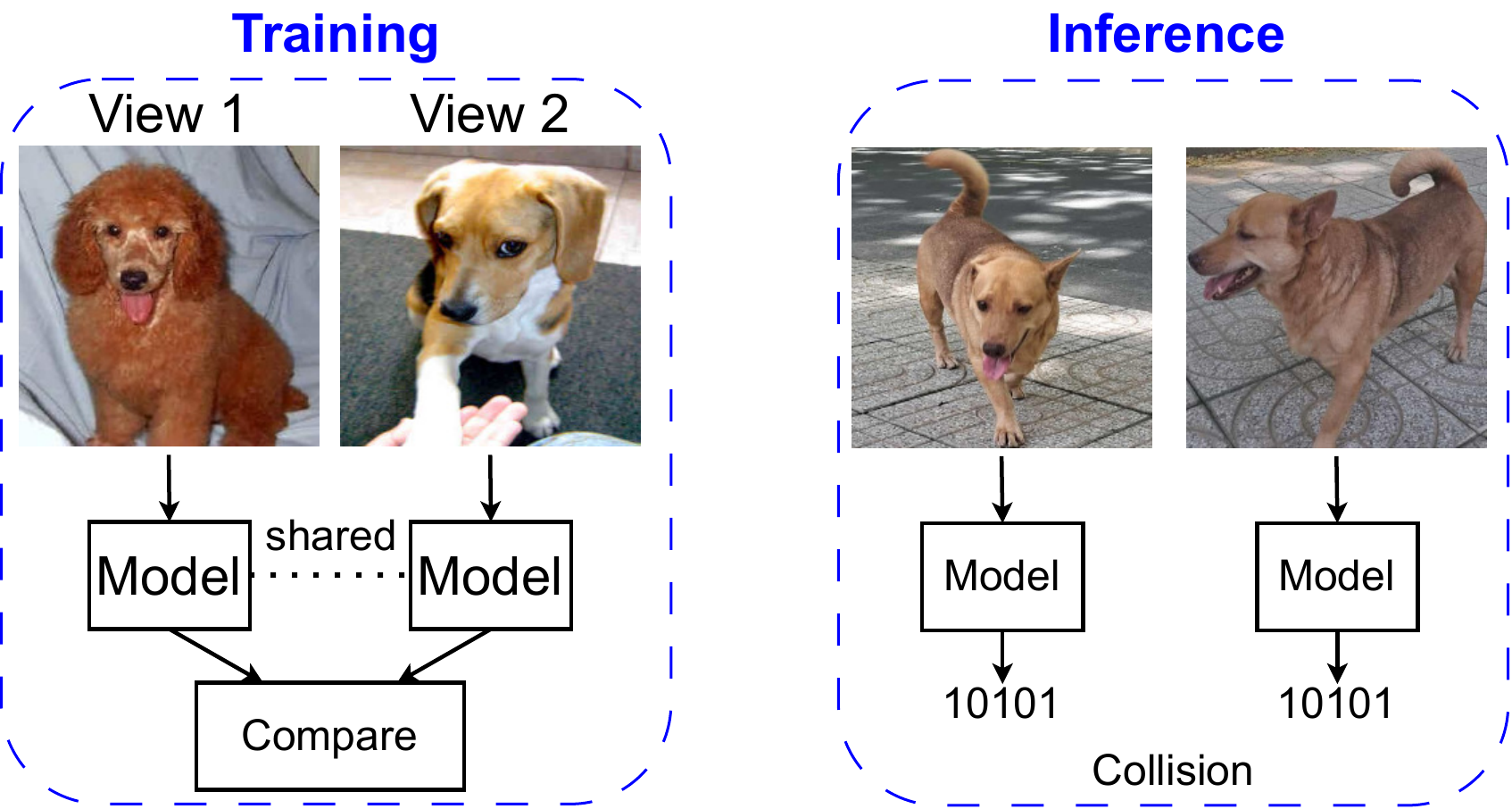}
        \end{minipage}
    }
    \\[0.3em]
    \subfloat[Our \acs{name}]{
        \begin{minipage}[b]{0.8\linewidth}
            \centering
            \includegraphics[width=\linewidth]{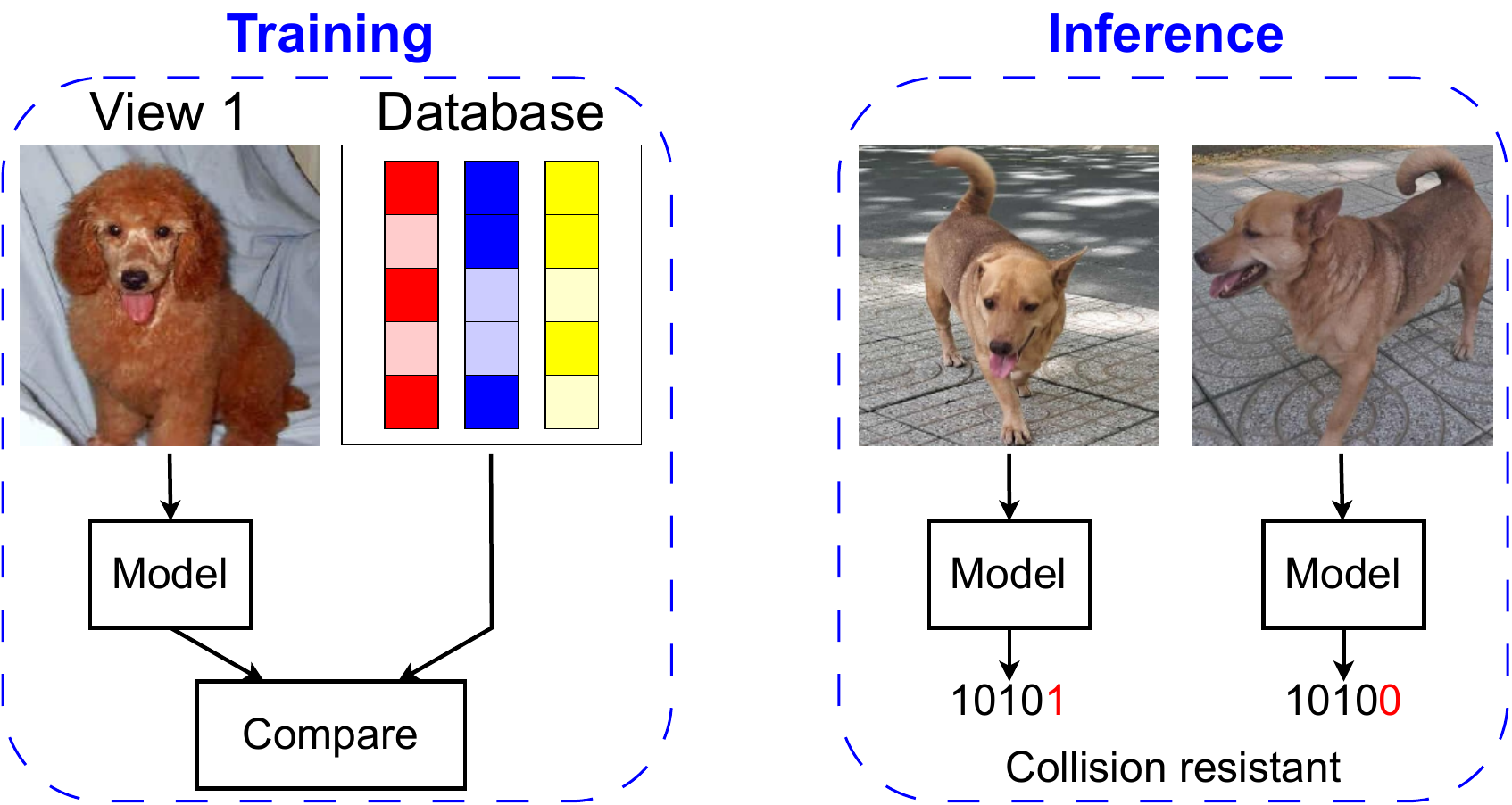}
        \end{minipage}
    }
    \caption{Comparison between existing multi-view contrastive methods and our proposed method.}
    \label{fig:intro}
\end{figure}

\Cref{fig:intro} contrasts our single-pass framework with existing multi-view contrastive methods. Unlike multi-view approaches that require multiple forward passes and pairwise comparisons between augmented views, our method performs instance discrimination through a memory-based formulation in a single forward pass. This design significantly reduces training overhead while directly optimizing similarity in the Hamming space, which is central to both collision resistance and retrieval efficiency.

\section{Experiment details}
\label{sec:setup}
We implemented our method in PyTorch~\cite{paszke2019pytorch} and conducted all experiments on a single NVIDIA H100 GPU. The models are trained for $100$ epochs with a batch size of $64$ using the Adam optimizer~\cite{kingma2014adam}. Detailed statistics of the datasets used in our experiments, including the number of training and test images and class distributions, are summarized in \cref{tab:dataset_details}.

\begin{table}[!htbp]
\caption{Summary of benchmark datasets used in our experiments.}
\label{tab:dataset_details}
\resizebox{\linewidth}{!}{
\begin{tabular}{c|cccccc}
\hline
Dataset & Flowers & CUB200 & Cars & Dogs & Food101 & NUS-WIDE \\ \hline
Train images & 2,040 & 5,994 & 8,144 & 12,000 & 75,750 & 157,043 \\
Test images & 6,149 & 5,794 & 8,041 & 8,580 & 25,250 & 2,100 \\
Classes & 102 & 200 & 196 & 120 & 101 & 21 \\ \hline
\end{tabular}
}
\end{table}

In \cref{tab:complexity_results} our method demonstrates the lowest training time and memory usage. As runtime is proportional to GFLOPs, our method is also faster in practice. On the Flowers dataset, our method takes 12 minutes for training, while 2-view methods (\eg~MeCoQ~\cite{wang2022contrastive}) take 22 minutes, and even higher for more view methods. For inference, our method achieves $1.61 \pm 0.0141$ ms per image, compared to $2.13 \pm 0.0101$ ms for codebook-based methods (\eg~MeCoQ~\cite{wang2022contrastive}).

\begin{figure}[h]
    \centering
    \includegraphics[width=0.6\linewidth]{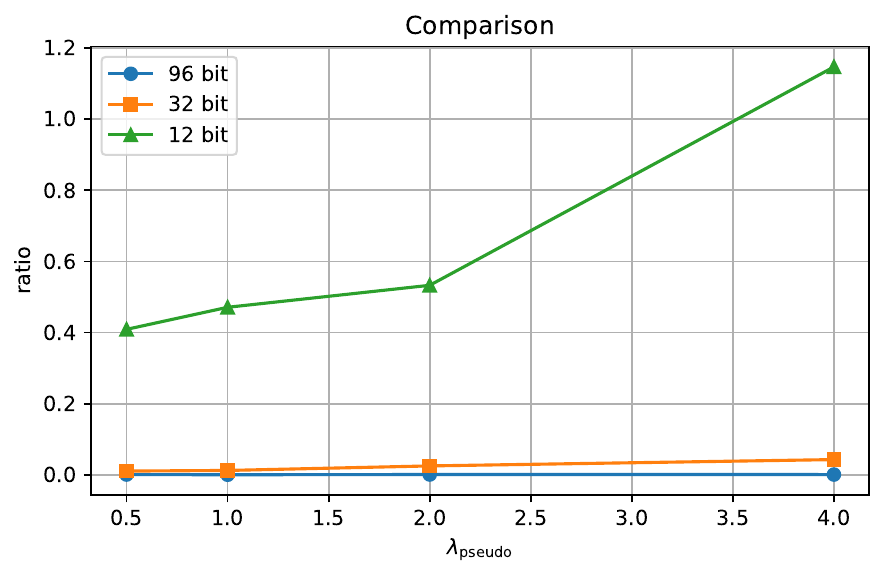}
    \caption{Ratio of collsions $\Pcollision$ (\textpertenthousand) vs. \ac{mAP} (\%) for different $\LambdaPseudo$ on the Flowers dataset. }
    \label{fig:ratio}
\end{figure}

In \cref{fig:ratio}, we analyze the relationship between the ratio of collisions $\Pcollision$ and the mean Average Precision (\ac{mAP}) for different values of $\LambdaPseudo$ on the Flowers dataset. As $\LambdaPseudo$ increases, we observe an increase in $\Pcollision$, especially for short code lengths. Although a larger $\LambdaPseudo$ can slightly improve retrieval performance, it introduces a substantially higher number of collisions, highlighting the trade-off between retrieval accuracy and collision resistance. This limitation becomes more critical at short code lengths, where the restricted hashing space makes the balance between retrieval performance and collision rate highly sensitive to $\LambdaPseudo$.

\section{Clustering-based extension}
\label{sec:codebook}

The previous \cref{sec:loss} introduces our single-pass approach for $\operatorname{sgn}$-based hashing. However, our framework is not restricted to this form, the single-pass strategy can also extend to clustering-based. Here, we present how our method adapts this idea to the codebook setting.

In the clustering-based hashing idea, the hash code of length $l$ is divided into $M$ smaller segments, or codebooks $\{C_{i}\}_{i=1}^M$. Each codebook holds up to $2^{\|C_i\|}$ codewords, where $\|C_i\|$ is the length of the $i$th codebook. Each codeword serves as the representative of a cluster within the corresponding codebook, so this design breaks the quantization task into multiple independent clustering subtasks. Following~\cite{wang2022contrastive}, if the hash bits are evenly split, the cluster weight matrix has the shape $[M, 2^{l/M}, D]$ where $D$ is the feature dimension. This results in $2^{l/M} \cdot M$ centroids in total. To keep it lightweight, we simply set $M = l$, meaning each bit is modeled by a separate codebook with only 2 centroids $\mu^{(C_k)}_{0, 1}$, reducing the total centroids to $2l$.

For clustering, we adopt the Deep Embedding Clustering (DEC) loss~\cite{xie2016unsupervised}, a widely used baseline in deep clustering. This choice not only keeps our method lightweight, but also allows for a fairer evaluation of the single-pass framework without external contributions from state-of-the-art clustering. Specifically, the total clustering loss is defined as:
\begin{equation}
    \LossCode =\frac{1}{M}\sum\limits_{i=1}^{M}{\LossDEC^{\left( C_i \right)}}.
\end{equation}
Here, the clustering loss can also inherently pull similar features closer together, as described in~\cite{xie2016unsupervised}.

We have revisited the core idea of codebook-based methods and now introduce our proposed clustering-based mechanism. For each codebook $C_k$, we compute the Euclidean distances $\delta^{(C_k)}_0 = \|z - \mu^{(C_k)}_0\|_2$ and $\delta^{(C_k)}_1 = \|z - \mu^{(C_k)}_1\|_2$, where $z$ is the concatenated vector of global and local with CSA. We then calculate the difference $\delta^{(C_k)} = \delta^{(C_k)}_0 - \delta^{(C_k)}_1$, which shows whether the feature is closer to the centroid, $0$ or $1$ (a positive value means closer to $\mu^{(C_k)}_1$ and vice versa). By concatenating these differences across all codebooks, we obtain a continuous vector $v' = [\delta^{(C_1)}, \delta^{(C_2)}, \ldots, \delta^{(C_l)}]$ which acts as an alternative representation analogous to $v$ in the $\tanh$ branch. This vector $v'$ acts as an alternative representation, analogous to $v$ in the $\tanh$ branch. We simply replace $v$ with $v'$ inside all loss computations. This allows the normalized Hamming distance loss, the pseudo labeling term, and the collision-sensitive attention module to operate seamlessly under the clustering-based formulation. Finally, the overall loss for the clustering-based variant becomes:

\begin{equation}
    L_{\text{clustering}} = \LossNHD + \LossPseudo + \LossAttention + \LossCode.
\end{equation}

At inference, the final hash code is obtained directly from the cluster assignments of each codebook:
\begin{equation}
    u_k =
    \begin{cases}
        1, & \text{if } \delta^{(C_k)}_1 < \delta^{(C_k)}_0, \\
        0, & \text{otherwise},
    \end{cases}
    \quad k = 1,\ldots,l,
\end{equation}
which yields the $l$-bit hash code $u = [u_1, u_2, \ldots, u_l]$.

\begin{table}[!htbp]
\caption{Retrieval \ac{mAP} (\%) of different configurations of clustering-based.}
\label{tab:cb_retrieval}
\resizebox{\linewidth}{!}{
\begin{tabular}{l|ccc|ccc}
\hline
\multirow{2}{*}{Setting}        & \multicolumn{3}{c|}{CUB200-2011}                 & \multicolumn{3}{c}{Oxford Flowers}               \\
                                & 12-bit         & 32-bit         & 96-bit         & 12-bit         & 32-bit         & 96-bit         \\ \hline
Clustering w/o pseudo, CSA, DEC & 2.48           & 8.85           & 17.36          & 3.55           & 30.94          & 48.62          \\
Clustering w/o CSA, DEC         & 18.71          & 24.68          & 35.19          & \textbf{36.88} & 49.15          & 54.23          \\
Clustering w/o DEC              & 17.64          & 27.59          & 36.73          & 35.58          & \textbf{50.86} & 59.47          \\
Clustering                      & \textbf{19.56} & \textbf{27.44} & \textbf{38.65} & 36.11          & 48.32          & \textbf{60.41} \\ \hline
\end{tabular}
}
\end{table}

\begin{table}[!htbp]
\caption{Collision probability $\Pcollision$ (\textpertenthousand) of different configurations of clustering-based.}
\label{tab:cb_collision}
\resizebox{\linewidth}{!}{
\begin{tabular}{l|ccc|ccc}
\hline
\multirow{2}{*}{Setting}        & \multicolumn{3}{c|}{CUB200-2011}             & \multicolumn{3}{c}{Oxford Flowers}           \\
                                & 12-bit        & 32-bit        & 96-bit       & 12-bit        & 32-bit        & 96-bit       \\ \hline
Clustering w/o pseudo, CSA, DEC & \textbf{2.42} & \textbf{0.11} & \textbf{0.0} & \textbf{2.86} & \textbf{0.10} & \textbf{0.0} \\
Clustering w/o CSA, DEC         & 10.85         & 5.26          & 0.43         & 16.37         & 8.18          & 0.68         \\
Clustering w/o DEC              & 8.42          & 1.79          & 0.19         & 12.07         & 1.46          & 0.31         \\
Clustering                      & 6.21          & 0.53          & \textbf{0.0} & 9.92          & 0.45          & \textbf{0.0} \\ \hline
\end{tabular}
}
\end{table}

As shown in \cref{tab:cb_retrieval} and \cref{tab:cb_collision}, the clustering-based extension achieves slightly lower retrieval performance but stronger collision resistance than the $\operatorname{sgn}$-based variant, at the cost of higher computational complexity. Consistent with the trends in \cref{tab:ablation_retrieval} and \cref{tab:ablation_collision}, using only $\LossNHD$ provides the best collision resistance but lower retrieval accuracy at shorter hash lengths (\eg~12 bits), while $\LossPseudo$ improves retrieval performance but yields more collisions.

\section{Mathematical proof}
\label{sec:proof}
In this section, we provide the mathematical analysis that formalizes the behavior of the proposed normalized Hamming distance loss. As discussed in \cref{sec:loss}, our objective is to directly optimize similarity in the Hamming space, ensuring that augmented views of the same instance remain close while representations of different instances are pushed toward the distance distribution of random binary vectors. The following derivations show that the loss inherently minimizes intra-instance distances and enforces inter-instance distances to concentrate around one, thereby explaining why the learned hash codes achieve strong collision resistance. 

We recall the definition of the normalized Hamming distance loss:
\begin{equation*}
    \LossNHD=-\log \frac{e^{s \frac{\left(2-d_{ii}\right)^2}{4} }}{ e^{s \frac{\left(2-d_{ii}\right)^2}{4}}+\sum\limits_{j=1,j\ne i}^{N}{e^{s{\left( 1-d_{ij} \right)}^2 }} },
\end{equation*}
where $\hat{v}_i=\tanh \left( \frac{s_1}{\left\| v_i \right\|_2}v_i \right)$, 
$\hat{W}_i=\tanh \left( \frac{s_1}{\left\| W_i \right\|_2} W_i \right)$, and $d_{ij} = \frac{\left\| \hat{v}_i-\hat{W}_j \right\|_1}{l} \in \left[0, 2 \right]$

Let
\begin{equation*}
    z^+ = s\frac{(2-d_{ii})^2}{4}, z_{j}^{-} = s(1-d_{ij})^2,
\end{equation*}
\begin{equation*}
    Z = e^{z^+} + \sum\limits_{j \ne i}{e^{z_{j}^{-}}}.
\end{equation*}
Then we can rewrite $\LossNHD = -\log \frac{e^{z^+}}{Z} = -z^+ + \log Z$

The gradient with respect to the positive pair $d_{ii}$ is:
\begin{equation*}
\begin{split}
        \frac{\partial \LossNHD}{\partial d_{ii}} &= - \frac{\partial z^+}{\partial d_{ii}} + \frac{1}{Z} \frac{\partial Z}{\partial d_{ii}} = - \frac{\partial z^+}{\partial d_{ii}} + \frac{e^{z^+}}{Z} \frac{\partial z^+}{\partial d_{ii}} \\ \quad
        &= - \left(1-\frac{e^{z^+}}{Z} \right)\frac{\partial z^+}{\partial d_{ii}} \\ \quad
        &= \left(1-\frac{e^{z^+}}{Z} \right) \frac{s}{2} \left( 2-d_{ii} \right).
\end{split}
\end{equation*}
Since $\frac{e^{z^+}}{Z} \in \left(0,1 \right)$ and $d_{ii} \in \left[0,2 \right]$, it follows that:
\begin{equation*}
    \frac{\partial \LossNHD}{\partial d_{ii}} = \left(1-\frac{e^{z^+}}{Z} \right) \frac{s}{2} \left( 2-d_{ii} \right) \ge 0.
\end{equation*}
Therefore, a descent step in gradient optimization decreases $d_{ii}$, forcing $\hat{v}_i$ and $\hat{W}_i$ closer.

\textbf{Corollary 1.} After optimization, for a small margin $\gamma^+ \approx 0$, we have:
\begin{equation*}
    d_{ii} = \frac{\left\|\hat{v}_i - \hat{W}_i\right\|_1}{l} \le \gamma^+, \forall i.
\end{equation*}

The normalized Hamming distance between $u_i$ and its augmented view $u'_i$ is:
\begin{equation*}
\begin{split}
    &E = \frac{1}{l} \left\| u_i - u'_i \right\|_1 \\ \quad
    &= \frac{1}{l} \left\| \left( u_i - \hat{v}_i \right) - \left(u'_i - \hat{v'}_i \right) + \left( \hat{v}_i - \hat{v'}_i \right) \right\|_1 \\ \quad
    &= \frac{1}{l} \left\| \left( u_i - \hat{v}_i \right) - \left(u'_i - \hat{v'}_i \right) + \left( \hat{v}_i - \hat{W}_i \right) - \left(  \hat{v'}_i - \hat{W}_i \right) \right\|_1
\end{split}
\end{equation*}
By the triangle inequality:
\begin{equation*}
    E \le d_{ii} + d'_{ii} +\epsilon_i +\epsilon'_i \le 2\gamma^+ + \epsilon_i +\epsilon'_i
\end{equation*}
where $\epsilon_i = \frac{\|u_i - \hat{v}_i\|_1}{l}$, $\epsilon'_i = \frac{\|u'_i - \hat{v'}_i\|_1}{l}$ is small when $s_1$ is large.

This condition ensures that augmented views of the same instance produce hash codes that remain close (and close to their reference $\hat{W}_i$), effectively guiding the model to learn consistent representations for retrieval.

The gradient with respect to the negative pair $d_{ij}, j \ne i$ is:
\begin{equation*}
\begin{split}
    \frac{\partial \LossNHD}{\partial d_{ii}} &= \frac{1}{Z}e^{z_{j}^-}\frac{\partial z_{j}^-}{\partial d_{ij}} \\ \quad
    &= \frac{e^{z_{j}^-}}{Z}2s \left(d_{ij}-1 \right).
\end{split}
\end{equation*}
If $d_{ij} \ge 1$, $\frac{\partial \LossNHD}{\partial d_{ii}} \ge 0$ descent step decreases $d_{ii}$. \\
If $d_{ij} < 1$, $\frac{\partial \LossNHD}{\partial d_{ii}} < 0$ descent step increases $d_{ii}$. \\
Therefore, gradient descent enforces $d_{ij}$ to approach $1$.

\textbf{Corollary 2.} After optimization, for a small margin $\gamma^- \approx 0$, we have:
\begin{equation*}
\begin{split}
    \left| d_{ij} - 1 \right| \le \gamma^-; \forall i,j: i \ne j \\ \quad
    1 - \gamma^- \le d_{ij} \le 1 + \gamma^-; \forall i,j: i \ne j.
\end{split}
\end{equation*}

We have the hash code $u_{i} = \operatorname{sgn} \left( v_{i} \right) = = \operatorname{sgn} \left( \hat{v}_i \right)$, the normalized Hamming distance between $u_i$ and $u_j$ is:
\begin{equation*}
\begin{split}
    &D = \frac{1}{l} \left\| u_i - u_j \right\|_1 \\ \quad
    &= \frac{1}{l} \left\| \left( u_i - \hat{v}_i \right) - \left(u_j - \hat{v}_j \right) + \left( \hat{v}_i - \hat{v}_j \right) \right\|_1 \\ \quad
    &= \frac{1}{l} \left\| \left( u_i - \hat{v}_i \right) - \left(u_j - \hat{v}_j \right) + \left( \hat{v}_i - \hat{W}_j \right) - \left(  \hat{v}_j - \hat{W}_j \right) \right\|_1
\end{split}
\end{equation*}
By the triangle inequality:
\begin{equation*}
    d_{ij} - d_{jj} - \epsilon_i -\epsilon_j \le D \le d_{ij} + d_{jj} +\epsilon_i +\epsilon_j
\end{equation*}
where $\epsilon_i = \frac{\|u_i - \hat{v}_i\|_1}{l}$ is small when $s_1$ is large.

Combining Corollaries 1 and 2, we obtain:
\begin{equation*}
\begin{split}
    D \ge d_{ij} - d_{jj} - \epsilon_i - \epsilon_j \ge  1 - \gamma^- - \gamma^+ - \epsilon_i - \epsilon_j \\ \quad
    D \le d_{ij} + d_{jj} + \epsilon_i + \epsilon_j \le  1 + \gamma^- + \gamma^+ + \epsilon_i + \epsilon_j
\end{split}
\end{equation*}
where $\gamma^-, \gamma^+, \epsilon_i, \epsilon_j$ are small and close to $0$ after optimization and $s_1$ is large.
So the Hamming distance $D$ between two hash codes is approximately $1$, ensuring that their spacing is similar to that between two random binary vectors ($D \approx 1$) and remain far from collisions ($D = 0$), thereby achieving collision resistance.

Therefore, $\LossNHD$ inherently minimizes the intra-instance distance $E$ and maintains inter-instance distances $D$ around one. A similar proof holds for $\LossPseudo$, which follows the same normalized Hamming distance formulation. Thus, the method jointly enforces collision resistance and learns semantically consistent retrieval representations directly in the Hamming space.

\section{Visualization}
\label{sec:vis}

\begin{figure*}[ht]
\centering
\newcommand{\ncols}{11}
\setlength{\tabcolsep}{0.25em}
\setlength{\fboxsep}{0em}
\setlength{\fboxrule}{0.12em}

\newlength{\imgwidthflower}
\setlength{\imgwidthflower}{\dimexpr(0.95\textwidth - \ncols\tabcolsep)/\ncols\relax}

\begin{tabular}{c|*{\numexpr\ncols-1\relax}{c}}
\centering
Query & 
\multicolumn{10}{c}{Top-10 retrieved images} \\

\includegraphics[width=\imgwidthflower]{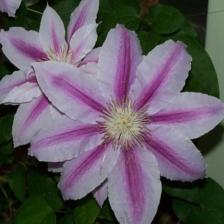} &
\includegraphics[width=\imgwidthflower]{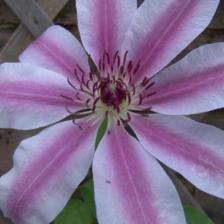} &
\includegraphics[width=\imgwidthflower]{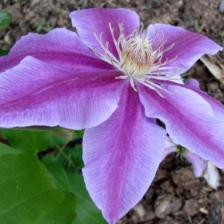} &
\includegraphics[width=\imgwidthflower]{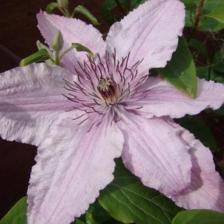} &
\includegraphics[width=\imgwidthflower]{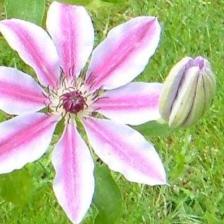} &
\includegraphics[width=\imgwidthflower]{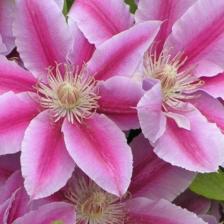} &
\includegraphics[width=\imgwidthflower]{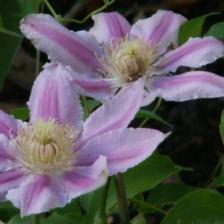} &
\includegraphics[width=\imgwidthflower]{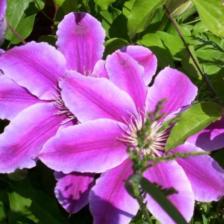} &
\includegraphics[width=\imgwidthflower]{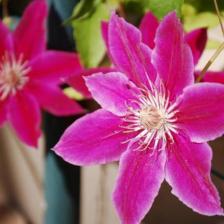} &
\includegraphics[width=\imgwidthflower]{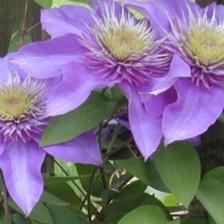} &
\includegraphics[width=\imgwidthflower]{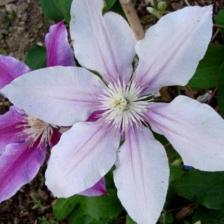} \\[-0.2em]

\includegraphics[width=\imgwidthflower]{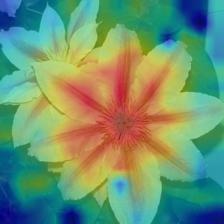} &
\includegraphics[width=\imgwidthflower]{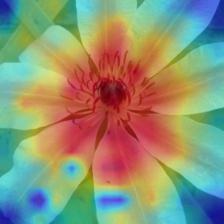} &
\includegraphics[width=\imgwidthflower]{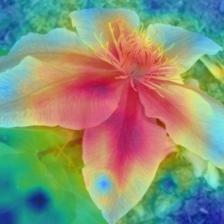} &
\includegraphics[width=\imgwidthflower]{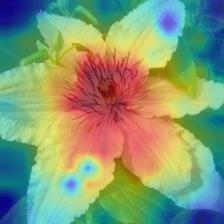} &
\includegraphics[width=\imgwidthflower]{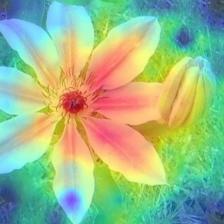} &
\includegraphics[width=\imgwidthflower]{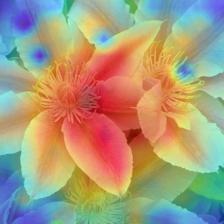} &
\includegraphics[width=\imgwidthflower]{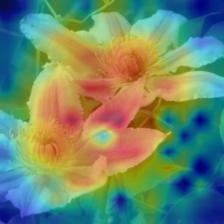} &
\includegraphics[width=\imgwidthflower]{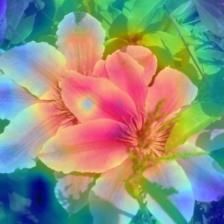} &
\includegraphics[width=\imgwidthflower]{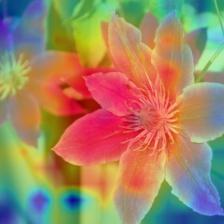} &
\includegraphics[width=\imgwidthflower]{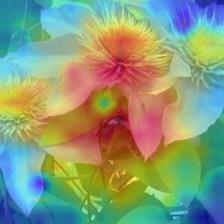} &
\includegraphics[width=\imgwidthflower]{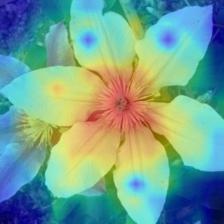} \\
\end{tabular}

\caption{Qualitative retrieval examples with 96-bit hash codes of \acs{name} with ViT backbone on Oxford Flowers dataset.}
\label{fig:flower}
\end{figure*}

\begin{figure*}[ht]
\centering
\newcommand{\ncols}{11}
\setlength{\tabcolsep}{0.25em}
\setlength{\fboxsep}{0em}
\setlength{\fboxrule}{0.12em}

\newlength{\imgwidthcub}
\setlength{\imgwidthcub}{\dimexpr(0.95\textwidth - \ncols\tabcolsep)/\ncols\relax}

\begin{tabular}{c|*{\numexpr\ncols-1\relax}{c}}
\centering
Query & 
\multicolumn{10}{c}{Top-10 retrieved images} \\

\includegraphics[width=\imgwidthcub]{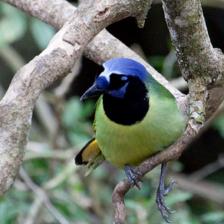} &
\includegraphics[width=\imgwidthcub]{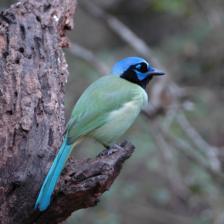} &
\includegraphics[width=\imgwidthcub]{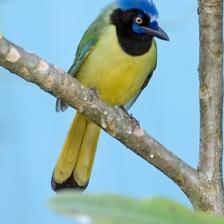} &
\includegraphics[width=\imgwidthcub]{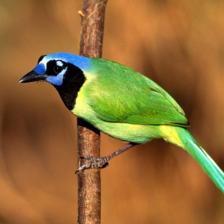} &
\includegraphics[width=\imgwidthcub]{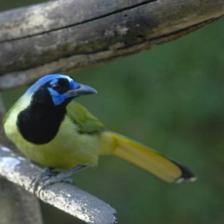} &
\includegraphics[width=\imgwidthcub]{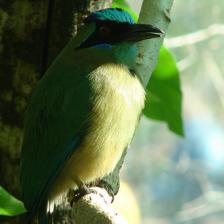} &
\includegraphics[width=\imgwidthcub]{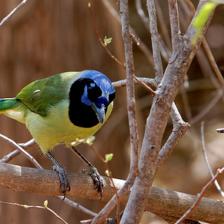} &
\includegraphics[width=\imgwidthcub]{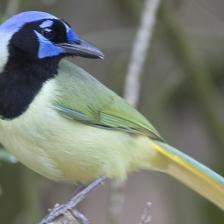} &
\includegraphics[width=\imgwidthcub]{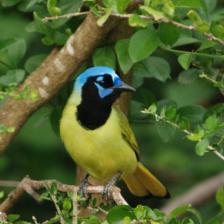} &
\includegraphics[width=\imgwidthcub]{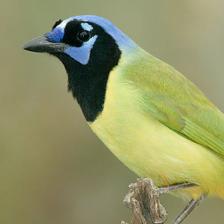} &
\includegraphics[width=\imgwidthcub]{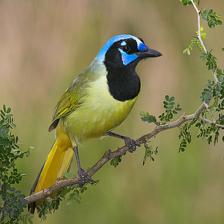} \\[-0.2em]

\includegraphics[width=\imgwidthcub]{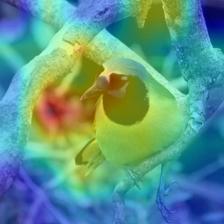} &
\includegraphics[width=\imgwidthcub]{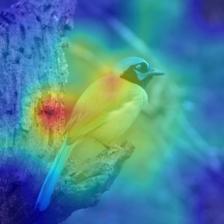} &
\includegraphics[width=\imgwidthcub]{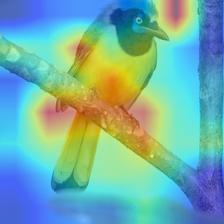} &
\includegraphics[width=\imgwidthcub]{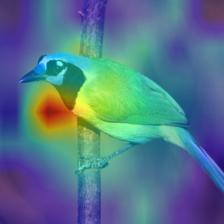} &
\includegraphics[width=\imgwidthcub]{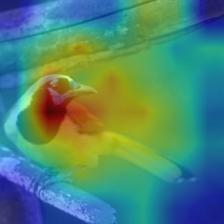} &
\includegraphics[width=\imgwidthcub]{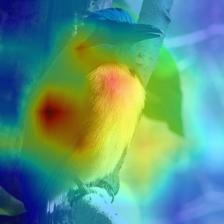} &
\includegraphics[width=\imgwidthcub]{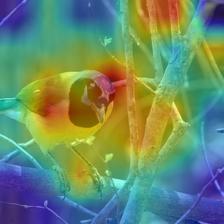} &
\includegraphics[width=\imgwidthcub]{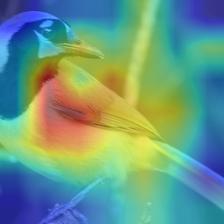} &
\includegraphics[width=\imgwidthcub]{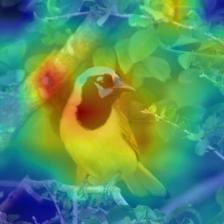} &
\includegraphics[width=\imgwidthcub]{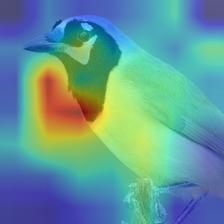} &
\includegraphics[width=\imgwidthcub]{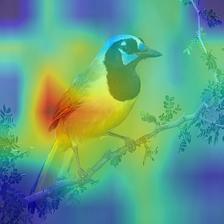} \\
\end{tabular}

\caption{Qualitative retrieval examples with 24-bit hash codes of \acs{name} with VGG16 backbone on CUB-200-2011 dataset.}
\label{fig:CUB}
\end{figure*}

\begin{figure*}[ht]
\centering
\newcommand{\ncols}{11}
\setlength{\tabcolsep}{0.25em}
\setlength{\fboxsep}{0em}
\setlength{\fboxrule}{0.12em}

\newlength{\imgwidthcar}
\setlength{\imgwidthcar}{\dimexpr(0.95\textwidth - \ncols\tabcolsep)/\ncols\relax}

\begin{tabular}{c|*{\numexpr\ncols-1\relax}{c}}
\centering
Query & 
\multicolumn{10}{c}{Top-10 retrieved images} \\

\includegraphics[width=\imgwidthcar]{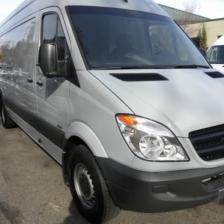} &
\includegraphics[width=\imgwidthcar]{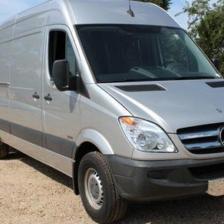} &
\includegraphics[width=\imgwidthcar]{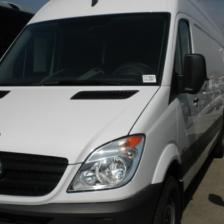} &
\includegraphics[width=\imgwidthcar]{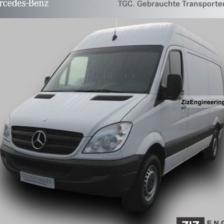} &
\includegraphics[width=\imgwidthcar]{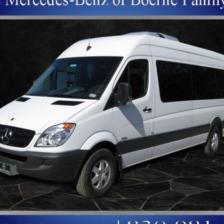} &
\includegraphics[width=\imgwidthcar]{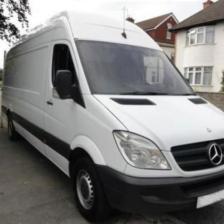} &
\fcolorbox{red}{white}{\includegraphics[width=\imgwidthcar]{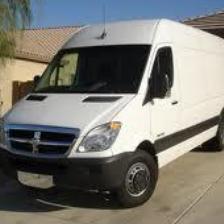}} &
\fcolorbox{red}{white}{\includegraphics[width=\imgwidthcar]{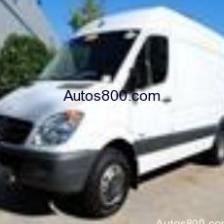}} &
\includegraphics[width=\imgwidthcar]{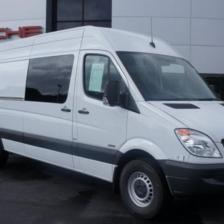} &
\includegraphics[width=\imgwidthcar]{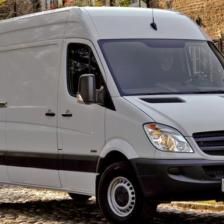} &
\fcolorbox{red}{white}{\includegraphics[width=\imgwidthcar]{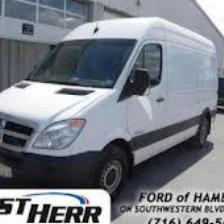}} \\[-0.2em]

\includegraphics[width=\imgwidthcar]{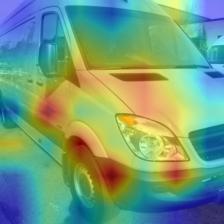} &
\includegraphics[width=\imgwidthcar]{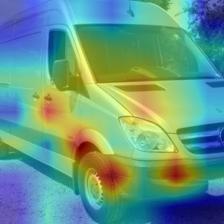} &
\includegraphics[width=\imgwidthcar]{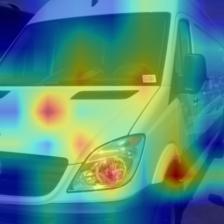} &
\includegraphics[width=\imgwidthcar]{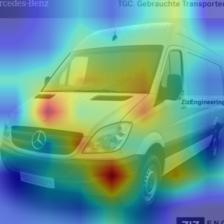} &
\includegraphics[width=\imgwidthcar]{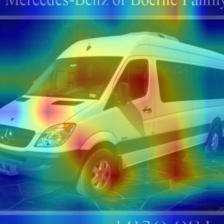} &
\includegraphics[width=\imgwidthcar]{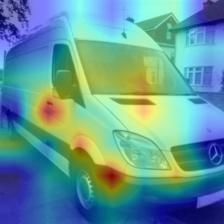} &
\fcolorbox{red}{white}{\includegraphics[width=\imgwidthcar]{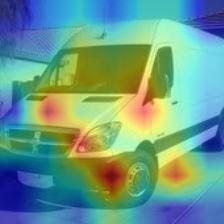}} &
\fcolorbox{red}{white}{\includegraphics[width=\imgwidthcar]{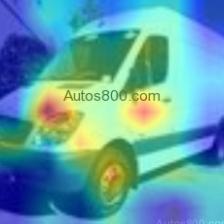}} &
\includegraphics[width=\imgwidthcar]{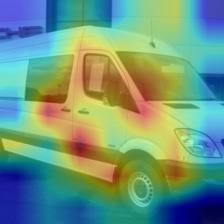} &
\includegraphics[width=\imgwidthcar]{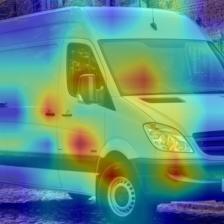} &
\fcolorbox{red}{white}{\includegraphics[width=\imgwidthcar]{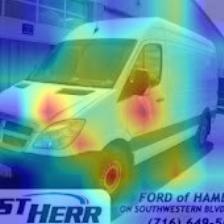}} \\
\end{tabular}

\caption{Qualitative retrieval examples with 48-bit hash codes of \acs{name} with ResNet backbone on Stanford Cars dataset.}
\label{fig:car}
\end{figure*}

\begin{figure*}[ht]
\centering
\newcommand{\ncols}{11}
\setlength{\tabcolsep}{0.25em}
\setlength{\fboxsep}{0em}
\setlength{\fboxrule}{0.12em}

\newlength{\imgwidthdog}
\setlength{\imgwidthdog}{\dimexpr(0.95\textwidth - \ncols\tabcolsep)/\ncols\relax}

\begin{tabular}{c|*{\numexpr\ncols-1\relax}{c}}
\centering
Query & 
\multicolumn{10}{c}{Top-10 retrieved images} \\

\includegraphics[width=\imgwidthdog]{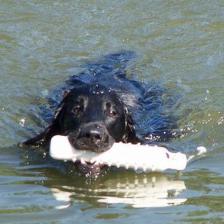} &
\fcolorbox{red}{white}{\includegraphics[width=\imgwidthdog]{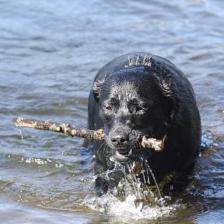}} &
\includegraphics[width=\imgwidthdog]{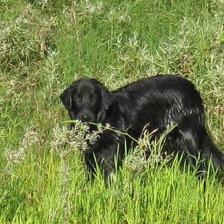} &
\includegraphics[width=\imgwidthdog]{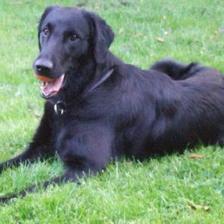} &
\includegraphics[width=\imgwidthdog]{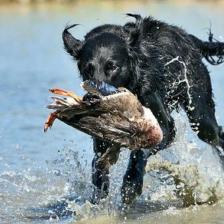} &
\includegraphics[width=\imgwidthdog]{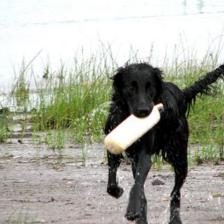} &
\includegraphics[width=\imgwidthdog]{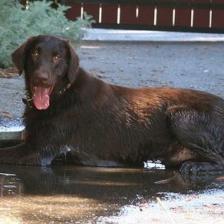} &
\includegraphics[width=\imgwidthdog]{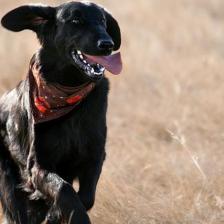} &
\includegraphics[width=\imgwidthdog]{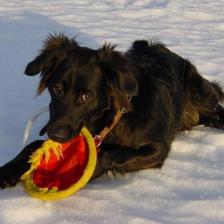} &
\includegraphics[width=\imgwidthdog]{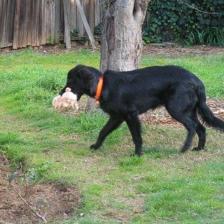} &
\includegraphics[width=\imgwidthdog]{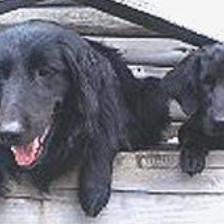} \\[-0.2em]

\includegraphics[width=\imgwidthdog]{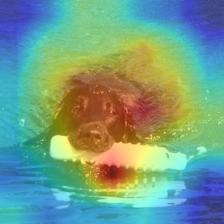} &
\fcolorbox{red}{white}{\includegraphics[width=\imgwidthdog]{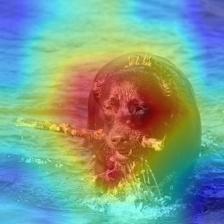}} &
\includegraphics[width=\imgwidthdog]{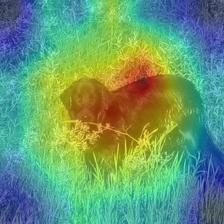} &
\includegraphics[width=\imgwidthdog]{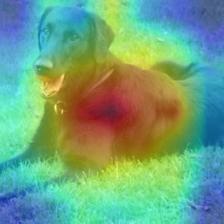} &
\includegraphics[width=\imgwidthdog]{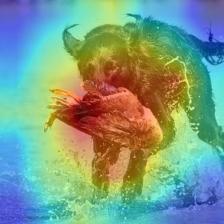} &
\includegraphics[width=\imgwidthdog]{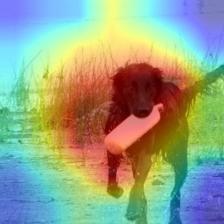} &
\includegraphics[width=\imgwidthdog]{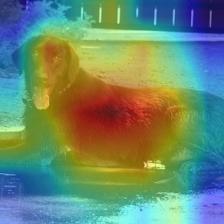} &
\includegraphics[width=\imgwidthdog]{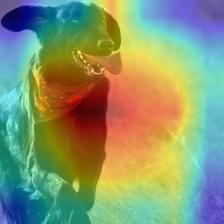} &
\includegraphics[width=\imgwidthdog]{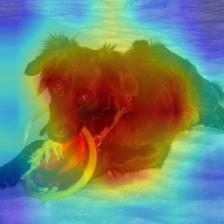} &
\includegraphics[width=\imgwidthdog]{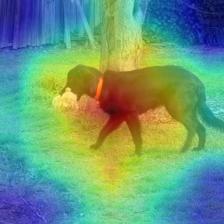} &
\includegraphics[width=\imgwidthdog]{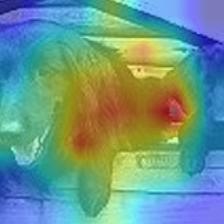} \\
\end{tabular}

\caption{Qualitative retrieval examples with 32-bit hash codes of \acs{name} with VGG16 backbone on Stanford Dogs dataset.}
\label{fig:dog}
\end{figure*}

\begin{figure*}[ht]
\centering
\newcommand{\ncols}{11}
\setlength{\tabcolsep}{0.25em}
\setlength{\fboxsep}{0em}
\setlength{\fboxrule}{0.12em}

\newlength{\imgwidthfood}
\setlength{\imgwidthfood}{\dimexpr(0.95\textwidth - \ncols\tabcolsep)/\ncols\relax}

\begin{tabular}{c|*{\numexpr\ncols-1\relax}{c}}
\centering
Query & 
\multicolumn{10}{c}{Top-10 retrieved images} \\

\includegraphics[width=\imgwidthfood]{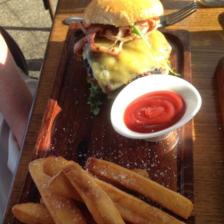} &
\fcolorbox{red}{white}{\includegraphics[width=\imgwidthfood]{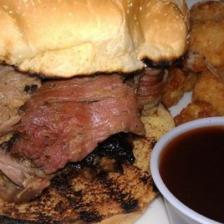}} &
\fcolorbox{red}{white}{\includegraphics[width=\imgwidthfood]{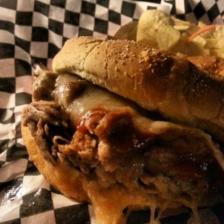}} &
\fcolorbox{red}{white}{\includegraphics[width=\imgwidthfood]{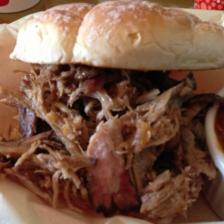}} &
\includegraphics[width=\imgwidthfood]{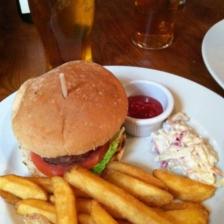} &
\includegraphics[width=\imgwidthfood]{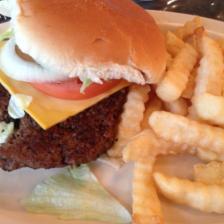} &
\fcolorbox{red}{white}{\includegraphics[width=\imgwidthfood]{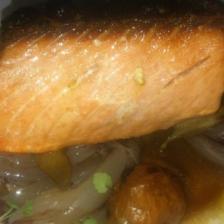}} &
\includegraphics[width=\imgwidthfood]{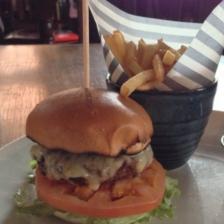} &
\includegraphics[width=\imgwidthfood]{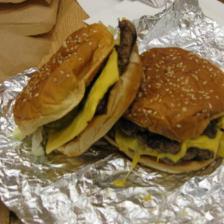} &
\includegraphics[width=\imgwidthfood]{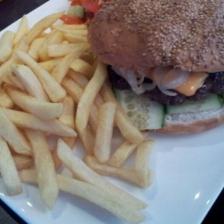} &
\fcolorbox{red}{white}{\includegraphics[width=\imgwidthfood]{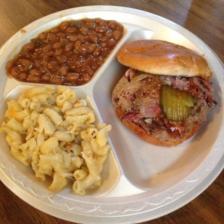}} \\
\end{tabular}

\caption{Qualitative retrieval examples with 12-bit hash codes of \acs{name} with ResNet backbone on Food-101 dataset.}
\label{fig:food}
\end{figure*}

\begin{figure*}[ht]
\centering
\newcommand{\ncols}{11}
\setlength{\tabcolsep}{0.25em}
\setlength{\fboxsep}{0em}
\setlength{\fboxrule}{0.12em}

\newlength{\imgwidthnus}
\setlength{\imgwidthnus}{\dimexpr(0.95\textwidth - \ncols\tabcolsep)/\ncols\relax}

\begin{tabular}{c|*{\numexpr\ncols-1\relax}{c}}
\centering
Query & 
\multicolumn{10}{c}{Top-10 retrieved images} \\

\includegraphics[width=\imgwidthnus]{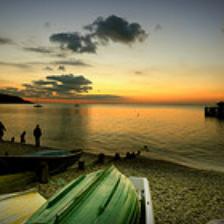} &
\includegraphics[width=\imgwidthnus]{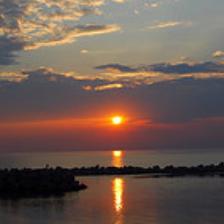} &
\includegraphics[width=\imgwidthnus]{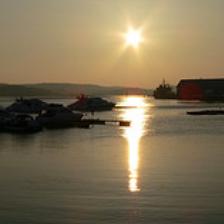} &
\includegraphics[width=\imgwidthnus]{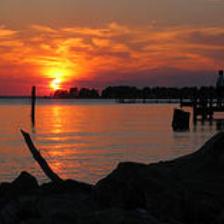} &
\includegraphics[width=\imgwidthnus]{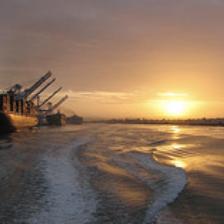} &
\includegraphics[width=\imgwidthnus]{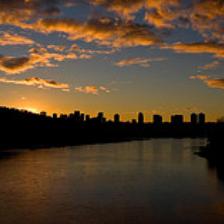} &
\includegraphics[width=\imgwidthnus]{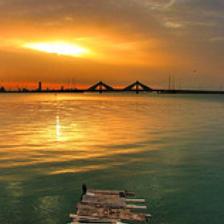} &
\includegraphics[width=\imgwidthnus]{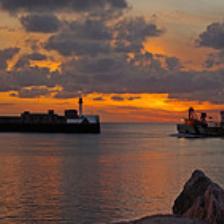} &
\includegraphics[width=\imgwidthnus]{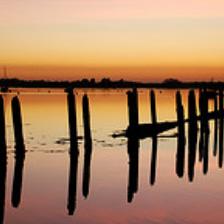} &
\includegraphics[width=\imgwidthnus]{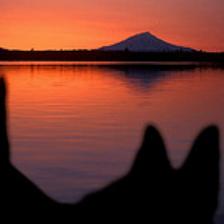} &
\includegraphics[width=\imgwidthnus]{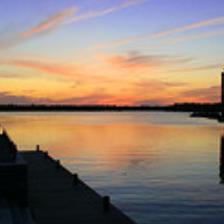} \\[-0.2em]

\includegraphics[width=\imgwidthnus]{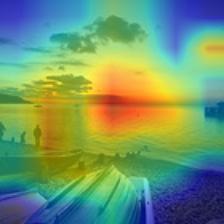} &
\includegraphics[width=\imgwidthnus]{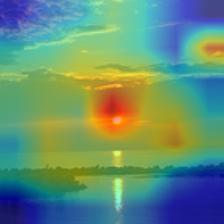} &
\includegraphics[width=\imgwidthnus]{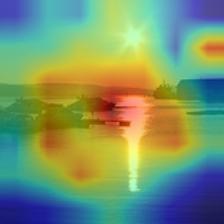} &
\includegraphics[width=\imgwidthnus]{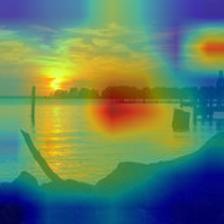} &
\includegraphics[width=\imgwidthnus]{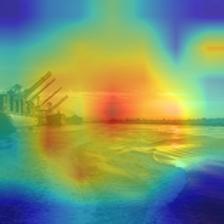} &
\includegraphics[width=\imgwidthnus]{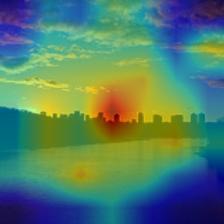} &
\includegraphics[width=\imgwidthnus]{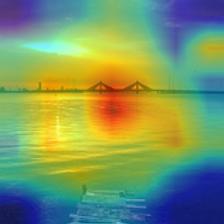} &
\includegraphics[width=\imgwidthnus]{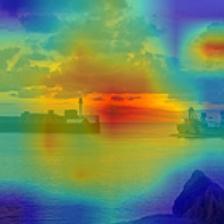} &
\includegraphics[width=\imgwidthnus]{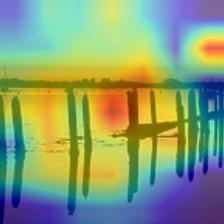} &
\includegraphics[width=\imgwidthnus]{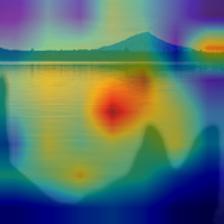} &
\includegraphics[width=\imgwidthnus]{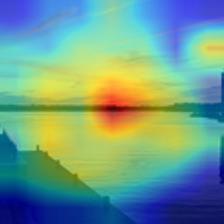} \\
\end{tabular}

\caption{Qualitative retrieval examples with 64-bit hash codes of \acs{name} with ResNet backbone on NUS-WIDE dataset.}
\label{fig:nus}
\end{figure*}

In this section, we provide additional qualitative retrieval examples across multiple datasets and backbones, as shown in \Cref{fig:flower,fig:CUB,fig:car,fig:dog,fig:food,fig:nus}. To ensure a comprehensive evaluation, we visualize random samples under different hash lengths ranging from $12$ to $96$ bits and across various architectures, including VGG16, ResNet50, and ViT-L/16. These examples cover a wide spectrum of retrieval difficulty, from low \acs{mAP} scenarios (\eg~8.44\% on Stanford Cars) to near-saturated performance (\eg~97.85\% on Oxford Flowers), as well as multi-label retrieval with high \acs{mAP}@5000 (\eg~85.0 on NUS-WIDE).

Across all visualizations, our method consistently produces collision-resistant and semantically coherent hash codes: retrieved images not only share the same fine-grained category but also exhibit similar local traits emphasized by the collision-sensitive attention. The attention maps typically highlight rare and discriminative cues such as textures, part-level color patterns, or object states. For instance, in \cref{fig:dog}, the model focuses on the specific action of a dog holding an object in its mouth, and the top retrieved images similarly depict dogs carrying objects. This behavior reflects the intended role of our attention module, prioritizing uncommon visual patterns that are crucial for avoiding collisions among visually similar samples.

In addition, although our method significantly improves collision resistance, a small number of collisions may still occur with very low probability. This phenomenon becomes especially visible in retrieval interfaces, where samples sharing identical hash codes are ranked first. Nonetheless, in our method these cases are rare and isolated. In contrast, existing hashing approaches with higher collision probability often retrieve many images with exactly the same binary code, making it difficult to distinguish visually similar and dissimilar samples. This not only harms interpretability but also affects retrieval performance, since multiple retrieved samples with equal Hamming distance collapse into the same rank. These observations further highlight the necessity of collision resistance in deep image hashing.

\begin{figure*}[ht]
\centering
\newcommand{\ncols}{11}
\setlength{\tabcolsep}{0.25em}
\setlength{\fboxsep}{0em}
\setlength{\fboxrule}{0.12em}

\newlength{\imgwidth}
\setlength{\imgwidth}{\dimexpr(0.9\textwidth - \ncols\tabcolsep)/\ncols\relax}

\begin{tabular}{c|*{\numexpr\ncols-1\relax}{c}}
\centering
Query & 
\multicolumn{10}{c}{Top-10 retrieved images} \\

\includegraphics[width=\imgwidth]{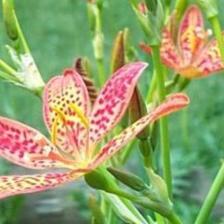} &
\includegraphics[width=\imgwidth]{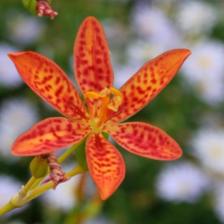} &
\includegraphics[width=\imgwidth]{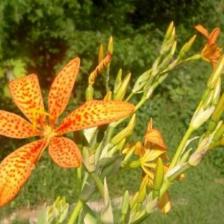} &
\includegraphics[width=\imgwidth]{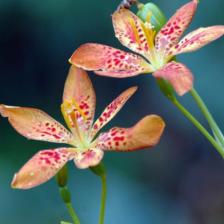} &
\includegraphics[width=\imgwidth]{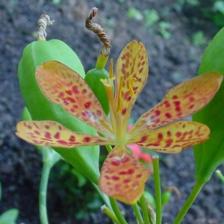} &
\includegraphics[width=\imgwidth]{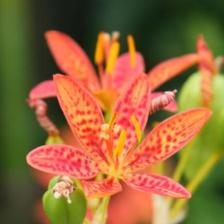} &
\includegraphics[width=\imgwidth]{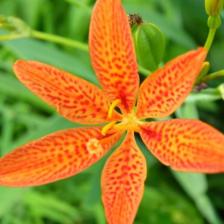} &
\includegraphics[width=\imgwidth]{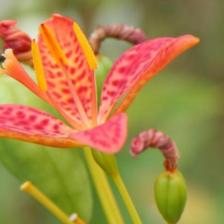} &
\includegraphics[width=\imgwidth]{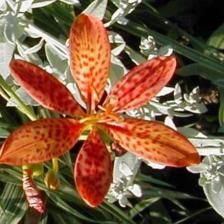} &
\includegraphics[width=\imgwidth]{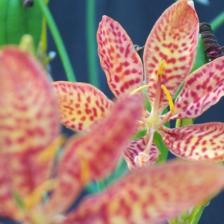} &
\includegraphics[width=\imgwidth]{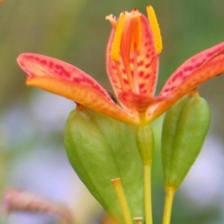} \\[-0.2em]

\includegraphics[width=\imgwidth]{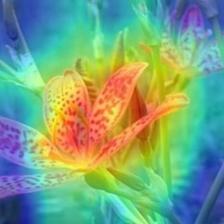} &
\includegraphics[width=\imgwidth]{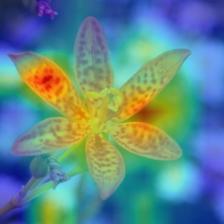} &
\includegraphics[width=\imgwidth]{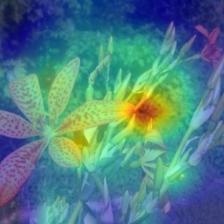} &
\includegraphics[width=\imgwidth]{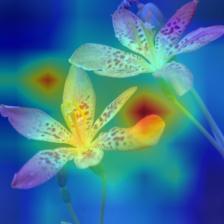} &
\includegraphics[width=\imgwidth]{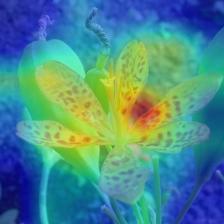} &
\includegraphics[width=\imgwidth]{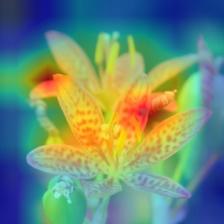} &
\includegraphics[width=\imgwidth]{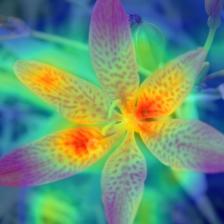} &
\includegraphics[width=\imgwidth]{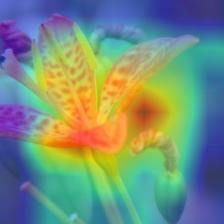} &
\includegraphics[width=\imgwidth]{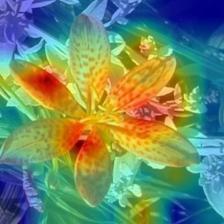} &
\includegraphics[width=\imgwidth]{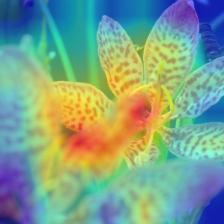} &
\includegraphics[width=\imgwidth]{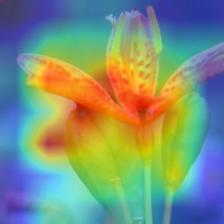} \\
\end{tabular}

\caption{Qualitative retrieval examples with 32-bit hash codes of \acs{name} with VGG16 backbone on Oxford Flowers and CUB-200-2011 datasets.}
\label{fig:qualitative}
\end{figure*}

\section{Limitation and discussion}
\label{sec:discussion}

While this work explicitly revisits collision resistance from cryptography and adapts it to deep image hashing, we note that cryptographic hashing and retrieval-oriented hashing serve different primary objectives. In retrieval, hashing functions act as a metric-preserving pre-filter, aiming to maintain similarity relationships in the Hamming space, whereas cryptographic hashing emphasizes security-related guarantees. Our motivation is not to equate these two settings, but to leverage collision resistance as a useful analytical lens for understanding and mitigating a practical issue in fine-grained retrieval: the frequent occurrence of multiple semantically different images sharing identical or indistinguishable hash codes. Such collisions can severely degrade retrieval ranking, lead to ambiguous query results, and underutilize the available hashing space when storing large-scale databases. From this perspective, explicitly modeling and measuring collision behavior provides complementary insight beyond retrieval accuracy alone.

Our framework prioritizes simplicity and efficiency through a single-pass design, which inevitably trades off some flexibility compared to multi-stage or multi-view approaches. Although the proposed normalized Hamming distance loss, memory-based instance discrimination, and collision-sensitive attention jointly improve collision resistance and retrieval performance, they are not claimed to be fundamentally orthogonal to all existing contrastive or clustering-based formulations. Rather, our contribution lies in unifying these components under direct Hamming space optimization and demonstrating that collision-aware design leads to more distinctive and robust hash codes in fine-grained scenarios. Future work may further analyze how collisions affect retrieval ambiguity and bit utilization, as well as investigate mechanisms for assigning clearer semantic roles to individual bits or groups of bits, enabling more interpretable and discriminative hashing representations without sacrificing efficiency.

\bibliographystyle{IEEEbib}
\bibliography{refs}

\end{document}